\pdfoutput=1

\documentclass{article}

\usepackage[main, final]{neurips_2025}

\usepackage{changepage} 
\newlength{\extramargin}
\usepackage[utf8]{inputenc}   
\usepackage[T1]{fontenc}      
\usepackage{microtype}        
\usepackage{inconsolata}      
\usepackage{times}            
\usepackage{pdflscape} 

\usepackage{amsmath}
\usepackage{amsfonts}
\usepackage{booktabs}         
\usepackage{multirow}
\usepackage{longtable}
\usepackage{rotating}
\usepackage{nicefrac}         
\usepackage{tablefootnote}
\usepackage{threeparttable}

\usepackage{graphicx}
\usepackage{subcaption}

\usepackage{hyperref}
\usepackage{url}
\usepackage{xcolor}

\newcommand{\xhdr}[1]{\paragraph{#1}}

\title{TiME: Tiny Monolingual Encoders for Efficient NLP Pipelines}

\author{%
  David Schulmeister \\
  EPFL \\
  \texttt{david.schulmeister@epfl.ch}
  \And
  Valentin Hartmann \\
  EPFL \\
  Timely Learning \\
  \texttt{valentin.hartmann@epfl.ch}
  \AND
  Lars Klein \\
  EPFL \\
  Timely Learning \\
  \texttt{lars.klein@epfl.ch}
  \And
  Robert West \\
  EPFL \\
  \texttt{robert.west@epfl.ch}
}

\begin{document}

\maketitle

\begin{abstract}
Today, a lot of research on language models is focused on large, general-purpose models. However, many NLP pipelines only require models with a well-defined, small set of capabilities. While large models are capable of performing the tasks of those smaller models, they are simply not fast enough to process large amounts of data or offer real-time responses. Furthermore, they often use unnecessarily large amounts of energy, leading to sustainability concerns and problems when deploying them on battery-powered devices.
In our work, we show how to train small models for such efficiency-critical applications. As opposed to many off-the-shelf NLP pipelines, our models use modern training techniques such as distillation, and offer support for low-resource languages. We call our models TiME (Tiny Monolingual Encoders) and comprehensively evaluate them on a range of common NLP tasks, observing an improved trade-off between benchmark performance on one hand, and throughput, latency and energy consumption on the other.\footnote{Models available at
\url{https://huggingface.co/collections/dschulmeist/time}, code at \url{https://github.com/epfl-dlab/TiME}.}
Along the way, we show that distilling monolingual models from multilingual teachers is possible, and likewise distilling models with absolute positional embeddings from teachers with relative positional embeddings.
\end{abstract}

\begin{figure*}[t]
    \centering
    \begin{subfigure}[b]{0.49\textwidth}
        \centering
        \includegraphics[width=\linewidth]{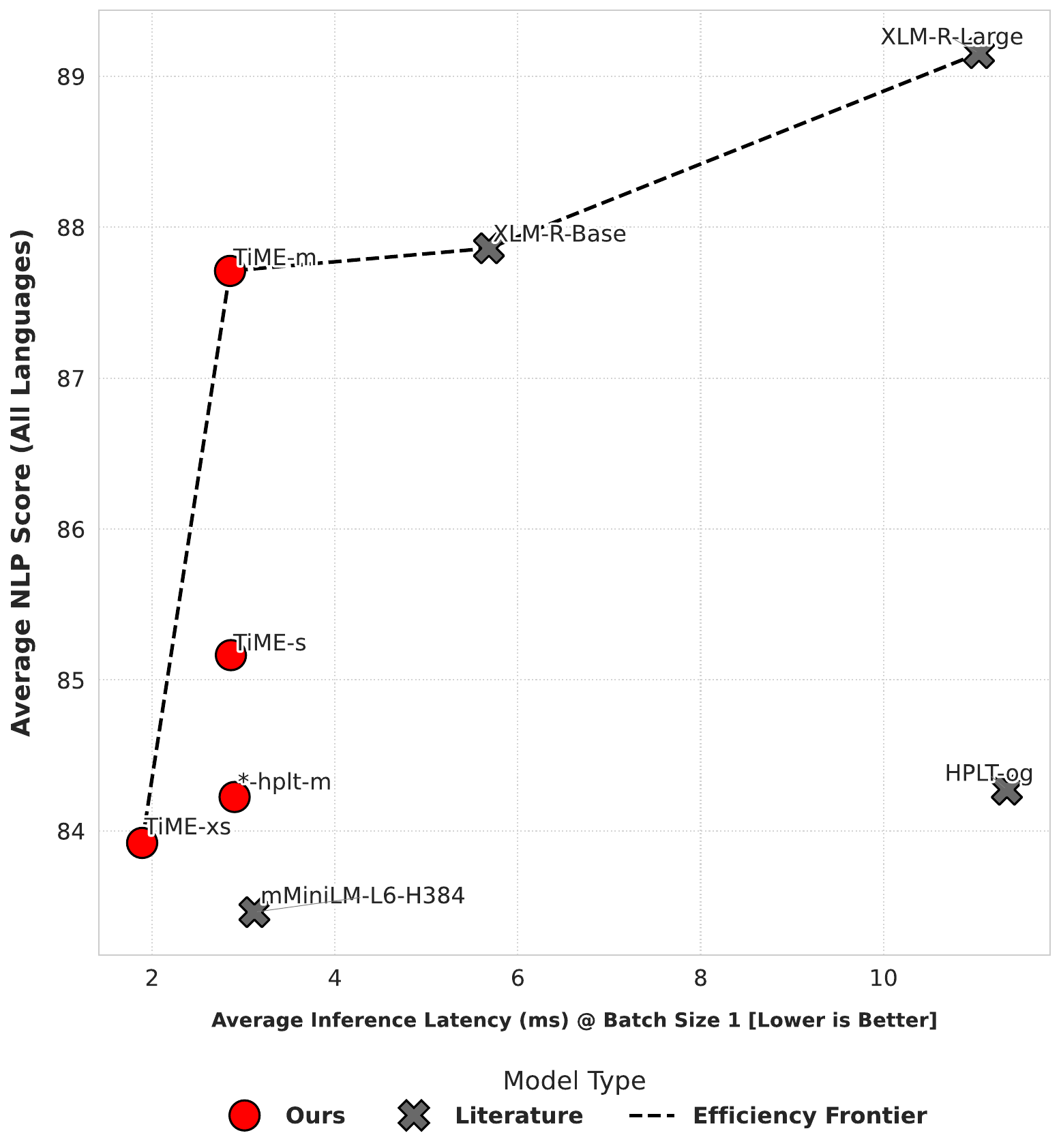}
        \caption{Performance vs. Latency}
        \label{fig:perf-vs-latency-avg}
    \end{subfigure}
    \hfill
    \begin{subfigure}[b]{0.49\textwidth}
        \centering
        \includegraphics[width=\linewidth]{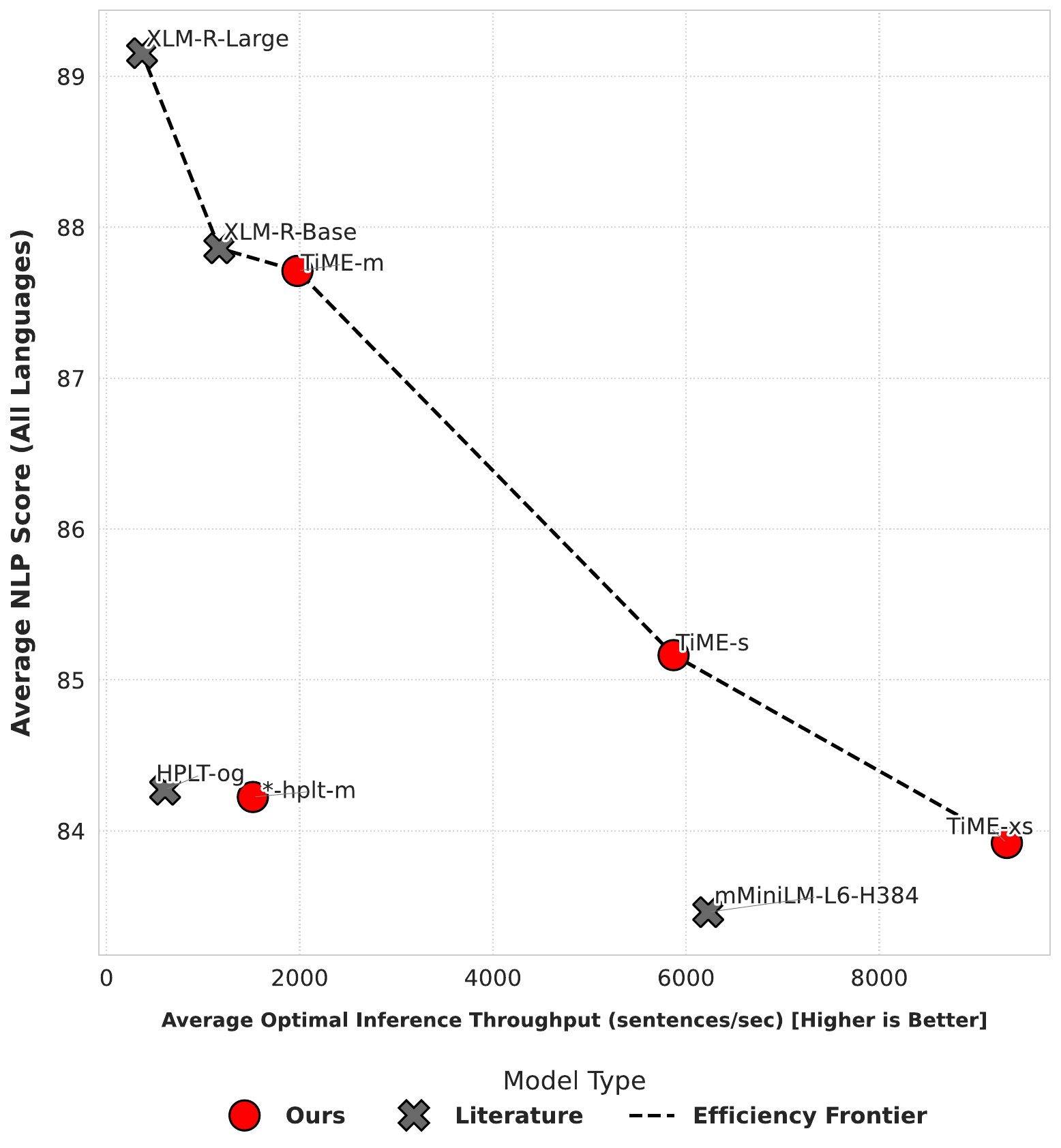}
        \caption{Performance vs. Throughput}
        \label{fig:perf-vs-throughput-avg}
    \end{subfigure}
    \caption{\textbf{Performance--efficiency trade-off, averaged across all languages.} Our distilled TiME models (red circles) are compared against baselines (grey crosses). The dashed line represents the efficiency frontier, connecting the models that offer the optimal trade-off. (a) plots the average NLP score against inference latency (ms), where lower is better. (b) plots the same score against throughput (samples/second) at the optimal batch-size, where higher is better. }
    \label{fig:performance-efficiency-tradeoff-avg}
\end{figure*}

\section{Introduction}

Transformer-based encoders such as BERT \cite{devlin2019bertpretrainingdeepbidirectional}, RoBERTa \cite{liu2019robertarobustlyoptimizedbert}, and XLM-RoBERTa (XLM-R) \cite{conneau2020unsupervisedcrosslingualrepresentationlearning} have become foundational to modern Natural Language Processing (NLP), achieving state-of-the-art results on a wide array of tasks. However, their substantial size, often comprising hundreds of millions or billions of parameters, and consequently high computational demands, pose significant challenges for deployment in time-critical or resource-constrained environments.

Support for non-English languages is often achieved by training multilingual models. While those offer versatility, their size can be prohibitive. Furthermore, their per-language capacity might be diluted compared to specialized monolingual models, which can offer optimal performance for individual languages \cite{Martin_2020, antoun2021araberttransformerbasedmodelarabic}. This creates a pressing need for efficient, yet high-performing, monolingual language models.

Knowledge Distillation (KD) \citep{hinton2015distillingknowledgeneuralnetwork} offers a path to compress large teacher models into smaller, more efficient students while retaining much of the original performance. We adopt the MiniLMv2 distillation framework \citep{wang2021minilmv2multiheadselfattentionrelation}, specifically its multi-head self-attention relation distillation, to create compact monolingual models. Our students are distilled from powerful teachers: the multilingual XLM-R-Large and strong monolingual models from the HPLT project \citep{aulamo-etal-2023-hplt, pyysalo2024hplt}.

We train models for 16 languages (full list in Table~\ref{table:appendix_detailed_scores_time}). For readability reasons, we focus on seven core languages in the main body of the paper and include results for the remaining languages in the appendix. The seven core languages were chosen to cover low-, medium-, and high-resource training data regimes: Irish (low); Urdu, Danish, Hungarian (medium); and English, German, French (high).

\xhdr{Contributions}

Our goal is to produce high-performing, yet significantly faster and more energy-efficient, monolingual encoder-only models that can be readily used for downstream applications. We call these models TiME (Tiny Monolingual Encoders). We make the following contributions:

\begin{itemize}
    \item We present a robust and practical MiniLMv2-based distillation pipeline and demonstrate its effectiveness in creating compact, high-performing TiME models for 16 languages, covering high- and low-resource regimes.
    \item We comprehensively evaluate these models on core NLP tasks (part-of-speech tagging, lemmatization, dependency parsing, named-entity recognition) and question answering, demonstrating competitive performance across all languages.
    \item Our distilled models achieve substantial inference speedups (up to 25$\times$) and energy efficiency improvement (up to 30$\times$) over their large teacher models and strong baselines (Table~\ref{table:main_summary_condensed_full}). Our evaluation focuses on this practical performance, measuring latency, throughput and energy use per sample to provide a more realistic assessment of efficiency than comparisons based on parameter count alone.
    \item We demonstrate successful knowledge transfer from teachers with relative position embeddings (LTG-BERT) to students with absolute position embeddings (standard BERT), and show that multilingual teachers can produce monolingual students rivaling those from specialized monolingual teachers.
    \item In the appendix, we include the results of additional experiments: NLP scores and speed for all 16 languages (Tables~\ref{table:appendix_detailed_scores_time} and \ref{tab:appendix_full_summary_16_langs}), an analysis of the trade-off between throughput and latency (Fig.~\ref{fig:latency-vs-throughput-batchsize}), and a comparison with the models used in the spaCy transformer pipelines (Sec.~\ref{ssec:spacy_comparison}).
\end{itemize}

\section{Related Work}
\label{sec:related_work}
The growing size of neural networks, especially Transformer-based ones \cite{vaswani2017attention}, has spurred significant research into model compression, where knowledge distillation (KD) from a large teacher model into a small student model \cite{hinton2015distillingknowledgeneuralnetwork} has become a powerful framework. Methods for distilling BERT-style models include DistilBERT \cite{sanh2020distilbertdistilledversionbert}, which uses a combination of losses on soft-target probabilities, and TinyBERT \cite{jiao2020tinybertdistillingbertnatural}, which leverages intermediate hidden states and attention matrices for a more fine-grained transfer. Other approaches, like MobileBERT \cite{sun2020mobilebertcompacttaskagnosticbert}, redesign the teacher and student architectures to facilitate layer-wise distillation. These methods showcase a range of strategies, primarily differing in the way of how knowledge is transferred from teacher to student.

Our work builds on the MiniLM family of distillation methods, which focus on transferring the internal mechanics of the self-attention mechanism. MiniLM \cite{wang2020minilmdeepselfattentiondistillation} introduced deep self-attention distillation, targeting the self-attention distributions and value-relations from the teacher's final layer. MiniLMv2 \cite{wang2021minilmv2multiheadselfattentionrelation}, the core method we employ, generalizes this by distilling fine-grained multi-head self-attention relations. These relations are defined as the scaled dot-products between pairs of query (Q), key (K), and value (V) vectors. This detailed approach crucially removes the constraint that student and teacher must have the same number of attention heads, allowing for greater flexibility in student architecture.

A key challenge in modern NLP is that large multilingual models like mBERT \cite{devlin2019bertpretrainingdeepbidirectional} and XLM-R \cite{conneau2020unsupervisedcrosslingualrepresentationlearning}, while enabling impressive cross-lingual transfer, often exhibit diluted per-language capacity and pose prohibitive deployment costs. This stands in contrast to specialized monolingual models, such as CamemBERT \cite{Martin_2020}, which can achieve superior performance. The approach of distilling compact monolingual students from large multilingual teachers is a promising strategy to combine the best of both worlds. \citet{electronics12041022} have previously demonstrated the viability of this strategy using a DistilBERT-style methodology, with a key contribution being a detailed analysis of vocabulary manipulation for low-resource languages.

Our work is complementary to these efforts but differs in several crucial aspects. First, we employ the more recent attention-relation transfer of MiniLMv2. Second, our analysis centers on practical performance--efficiency trade-offs. Finally, we demonstrate the robustness of our approach by successfully bridging architectural mismatches, such as distilling knowledge from a teacher with relative position embeddings into a student with standard absolute position embeddings.

\section{Methodology}

We replicate the distillation setup of the MiniLMv2 models \cite{wang2021minilmv2multiheadselfattentionrelation} and use it to train efficient monolingual models from mono- and multilingual teachers. We evaluate the models' performance when fine-tuned on typical NLP pipeline tasks, and the speedup over their larger teachers.

\subsection{Distillation}

We adopt the multi-head self-attention relation distillation strategy from MiniLMv2 \cite{wang2021minilmv2multiheadselfattentionrelation}. The core idea is to train the student model to mimic the self-attention relations of a specific teacher layer. These relations are computed as the scaled dot-product of pairs of query (Q), key (K), and value (V) vectors. The total loss is a weighted sum of KL-divergence losses between teacher and student relations for chosen pairs:
\begin{equation*}
\mathcal{L}_{\text{Distill}} = \sum_{(m,n) \in \mathcal{R}} w_{mn} D_{KL}(Rel_T^{mn} || Rel_S^{mn}),
\end{equation*}
where $\mathcal{R}$ is the set of chosen relation pairs, $w_{mn}$ are their weights, and $Rel_T^{mn}$ and $Rel_S^{mn}$ are the attention relations (distributions over sequence positions after softmax) for the teacher and student, respectively. Following \cite{wang2021minilmv2multiheadselfattentionrelation} we use Q-Q, K-K, and V-V relations with equal weight. The raw dot-products $A_L^{mn} = \text{vector}_m \cdot \text{vector}_n^T / \sqrt{d_k}$ (where $d_k$ is the dimension of the key/query vectors) are used to compute these relations, with a softmax function applied before calculating the KL divergence.

\subsection{Models}

\xhdr{Teachers} As the multilingual teacher we use the XLM-R-Large model \cite{conneau2020unsupervisedcrosslingualrepresentationlearning}, and as monolingual teachers the models from the HPLT project \cite{aulamo-etal-2023-hplt, pyysalo2024hplt}.

\xhdr{Students}
The student models are Transformer encoders with varying depths and widths. We define three sizes:
\begin{itemize}
\item \textbf{Medium (m):} 6 layers, 768 hidden size (\(L_S = 6,\ H_S = 768\))
\item \textbf{Small (s):} 6 layers, 384 hidden size (\(L_S = 6,\ H_S = 384\))
\item \textbf{Extra-Small (xs):} 4 layers, 384 hidden size (\(L_S = 4,\ H_S = 384\))
\end{itemize}
For all students, the intermediate feed-forward size is \(4 \times H_S\), and the number of attention heads is 12. The student and the teacher always share the same tokenizer.

Our choice of layer 19 for the XLM-R-Large teacher directly follows the recommendation in the original MiniLMv2 paper \cite{wang2021minilmv2multiheadselfattentionrelation}, which empirically found this layer to be the most effective knowledge source for large RoBERTa-style architectures like XLM-R.
The number of relation heads ($A_r$) is set to 64 for XLM-R-Large and 48 for the other teachers.

It is worth noting that the HPLT teacher models \cite{pyysalo2024hplt} use the LTG-BERT architecture \cite{samuel-etal-2023-ltgbert}, which incorporates modifications such as GeGLU activations and relative position embeddings. We deliberately chose to distill into a standard BERT architecture with absolute position embeddings. This decision was motivated by two factors: First, standard BERT architectures are broadly compatible with existing NLP tooling, ensuring our models are easy to adopt. Second, we observed that the LTG-BERT architecture can be substantially slower in practice. We found that the MiniLMv2 distillation method is robust enough to bridge these architectural differences, successfully transferring knowledge from the teacher despite the change in position embedding strategy and activations.

\xhdr{Training}
\label{ssec:training_setup}
We train models using the AdamW optimizer with $\beta_1=0.9$ and $\epsilon=1\mathrm{e}{-}6$. We set $\beta_2=0.98$ for distillation from XLM-R-Large and $\beta_2=0.999$ for HPLT models, following the original MiniLMv2 hyperparameter tuning which found different optimal values for RoBERTa-style and BERT-style teachers, respectively. The learning rate is $5.5\mathrm{e}{-}4$, with a 4 000-step linear warmup and subsequent linear decay. We train each model for 200,000 steps on NVIDIA A100, H100, and H200 GPUs with BF16 mixed-precision and an effective batch size of 256, saving a checkpoint every 10,000 steps. In Sec.~\ref{sec:results}, we report the results for the best checkpoints.

\subsection{Checkpoint Selection}
\label{ssec:model_selection}

We select the optimal checkpoint from the 20 checkpoints saved during each 200,000-step training run. For each language, we evaluate all intermediate checkpoints and keep the one that minimizes the MiniLMv2 distillation loss on an external validation set that is not seen during pre-training:

\begin{itemize}
  \item \textbf{Irish (ga)} – we use the dev split of the
        \textsc{FLORES-200} multilingual benchmark
        \citep{goyal2022flores101, flores_hf}.\footnote{Dataset ID
        \texttt{facebook/flores} on Hugging Face.}
  \item \textbf{All other languages} – we use the source/target side of
        the \textsc{WMT24++} parallel corpus \citep{wmt24pp, wmt24pp_arxiv}
        that corresponds to the language pair
        \texttt{en}$\leftrightarrow$\texttt{XX}.  For English we take the
        English source side, for the remaining languages the respective
        target side.
\end{itemize}

Note that for Irish we had to find a different validation set, since Irish is not contained in the WMT24++ dataset. 
This checkpoint selection strategy is particularly important. While longer distillation can sometimes yield further improvements \citep{wang2021minilmv2multiheadselfattentionrelation}, it also significantly increases computational cost. More critically, for low-resource languages where the unique training data is limited, extended training (equivalent to many epochs over these smaller datasets) heightens the risk of the student model overfitting to the distillation data. By evaluating on an external validation set unseen during distillation, we want to identify checkpoints that generalize well and strike a balance between effective knowledge transfer and robustness, especially for these resource-scarce scenarios. 

\subsection{Datasets and Evaluation Tasks}

\paragraph{Training data and languages.}
All student models are distilled using language-specific subsets of the CulturaX dataset \cite{nguyen2023culturaxcleanedenormousmultilingual}.
We train models for high-resource languages (English, en; German, de; French, fr), medium-resource languages (Danish, da; Hungarian, hu; Urdu, ur) and a low-resource language (Irish, ga).
For English, we only train a single model, meant as a reproduction of the MiniLMv2 paper \cite{wang2021minilmv2multiheadselfattentionrelation}, since their focus is on training English models.
To have a reference for a small multilingual model that supports all of those languages simultaneously, we compare with mMiniLM-L6-H384, distilled from XLM-R-Large \cite{wang2021minilmv2multiheadselfattentionrelation}. In the appendix, we report results on models for nine additional languages that were trained in the same way.

\paragraph{NLP tasks.}
We evaluate the models on a suite of common NLP tasks: named entity recognition (balanced F1 score; NER), part-of-speech tagging (accuracy; AllTags), lemmatization (accuracy; Lemma) and dependency parsing (labeled attachment score; LAS). We adopt the evaluation framework used for the HPLT models \cite{pyysalo2024hplt}. This involves benchmarking on relevant treebanks from Universal Dependencies \cite{de2021universal} for POS tagging, lemmatization, and parsing, and on the WikiAnn dataset \cite{rahimi-etal-2019-massively} for NER. Detailed results for all tasks and languages are presented in Table~\ref{table:appendix_detailed_scores_time} in the appendix.

\paragraph{Speed.}
For measuring inference throughput and latency, we sample 110 batches of 32 sentences from the CulturaX subsets for the different languages. We use 10 batches for warmup, and then measure wall-clock time for the remaining 100 batches. The benchmarks are run on an NVIDIA A100-SXM4-80GB GPU.

\paragraph{Energy consumption.}
GPU energy consumption is recorded using the same inference setup as for measuring latency and throughput by polling \texttt{nvidia-smi}. During each post-warmup window, we sample
\texttt{nvidia-smi --query-gpu=power.draw --format=csv,noheader,nounits}
at 10\,Hz and average over the entire window.

\paragraph{Question answering.}
To test whether besides NLP tasks our models are also capable of tasks that require more general knowledge, we evaluate the English and German models on the MLQA question answering benchmark \cite{lewis-etal-2020-mlqa} (only available for en and de).

\section{Experiments and Results} 
\label{sec:results}

This section details the performance of our distilled student models against teacher models and other relevant baselines. All results correspond to the best-performing checkpoint for each configuration (selected as described in Section~\ref{ssec:model_selection}). Our distilled models, which we call TiME (Tiny Monolingual Encoders), follow the naming convention \texttt{TiME-\{lang\}-\{size\}} (when distilled from XLM-R-Large). Size codes denote the architectures defined in Section 3.2: `xs`, `s`, and `m`. We re-evaluated all baselines, including the original HPLT original models (\texttt{\{lang\}-hplt-og}). Unless noted, figures aggregate over a seven-language set intentionally spanning low/medium/high resource tiers (ga; ur/da/hu; en/de/fr), to ensure conclusions hold across resource levels.

\subsection{Performance on Core NLP Tasks}
Table~\ref{table:main_summary_condensed_full} summarizes the performance on core NLP tasks. The \texttt{TiME-m} models, our best-performing students, retain 98.4\% of the average score of the `XLM-R-Large` teacher (Table~\ref{table:main_summary_condensed_full}). This is achieved with a 58\% reduction in parameter count (236M vs. 560M).

\begin{table}[t]
\centering
\small
\setlength{\tabcolsep}{2pt}
\renewcommand{\arraystretch}{1.4}

\begin{threeparttable}
\begin{tabular}{@{}l c c ccccccc c c c c c@{}}
\toprule
Model & \#P (M) & \#L & da & de & en & fr & ga & hu & ur &
\textbf{Avg} &
\shortstack{Lat.\\ Impr. \\($\times$)} &
\shortstack{T-put\\ Impr. \\($\times$)} &
\shortstack{Optim\\ J/sample} &
\shortstack{J/sample\\ Impr. \\($\times$)} \\
\midrule
\multicolumn{15}{@{}l}{\textit{Baselines}} \\
HPLT (our eval)     & 150 & 12 & 88.8 & 89.4 & 91.1 & 93.7 & 80.3 & 70.9 & 75.6 & 84.3 & 1.0 & 1.6 & 0.61 & 1.3$\times$ \\
XLM-R-Base          & 278 & 12 & 91.0 & 89.5 & 91.5 & 94.2 & 81.0 & 80.0 & 87.9 & 87.9 & 1.9 & 3.2 & 0.25 & 3.3$\times$ \\
mMiniLM\tnote{a}    & 107 & 6  & 86.1 & 84.4 & 88.5 & 91.4 & 70.2 & 75.1 & 85.3 & 83.5 & \textbf{3.5} & \textbf{16.9} & 0.04 & 19.4$\times$ \\
XLM-R-Large         & 560 & 24 & \textbf{92.4} & \textbf{90.5} & \textbf{92.4} & \textbf{95.1} & \textbf{83.1} & \textbf{81.6} & \textbf{89.0} & \textbf{89.2} & 1.0 & 1.0 & 0.82 & 1.0$\times$ \\
\cmidrule(l){1-15}
\multicolumn{15}{@{}l}{\textit{Our Students}} \\
\textbf{TiME-m}     & \textbf{236} & \textbf{6} & \textbf{89.9} & \textbf{88.4} & \textbf{91.1} & \textbf{93.2} & \textbf{82.4} & \textbf{80.8} & \textbf{88.1} & \textbf{87.7} & 3.9 & 5.4  & 0.12 & 6.6$\times$ \\
TiME-s              & 107 & 6  & 88.1 & 86.8 & 88.7 & 91.9 & 77.5 & 76.9 & 86.3 & 85.2 & 3.9 & 15.9 & 0.04 & 18.7$\times$ \\
TiME-xs             & 103 & 4  & 86.2 & 84.8 & 87.6 & 91.1 & 75.8 & 76.0 & 86.0 & 83.9 & \textbf{5.8} & \textbf{25.2} & 0.02 & 30.2$\times$ \\
*-hplt-m            & 69  & 6  & 88.1 & 87.0 & 90.0 & 91.8 & 80.2 & 77.8 & 74.5 & 84.2 & 3.8 & 4.1  & 0.18 & 4.5$\times$ \\
\bottomrule
\end{tabular}

\begin{tablenotes}
\footnotesize
\item[a] L6-H384 version.
\end{tablenotes}
\vspace{8pt}
\caption{\textbf{Average NLP task scores and efficiency metrics for all student models and baselines.} The final `Avg` column is the macro-average score across languages. Latency and throughput are shown as relative speedup factors ($\times$), computed against XLM-R-Large (set to $1.0\times$). For latency, higher means faster (e.g., $2.0\times$ means half the latency of XLM-R-Large). For throughput, higher means more sentences/sec. Throughput is measured at the optimal batch size for each model. Best student and best overall scores per language are \textbf{bolded}.}
\label{table:main_summary_condensed_full}
\end{threeparttable}

\end{table}

Our approach shows strong performance in varied settings, including the low-resource language Irish (ga) and the morphologically complex Hungarian (hu), where the models recover over 99\% of the teacher's score. Our distilled \texttt{TiME-m} models consistently outperform the `mMiniLM-L6-H384` baseline across all languages (Table~\ref{table:main_summary_condensed_full}). They achieve performance comparable to the larger XLM-R-Base model (278M parameters) while being approximately 15\% smaller (236M parameters).

While the overall parameter savings over XLM-R-Base appear modest (15\%), this is misleading: A large fraction of the total parameters is in the linear embedding layer that is shared with the teacher. The reduction in the actual Transformer layers, the part responsible for most inference time, is substantially larger. This leads to greater real-world efficiency gains than the parameter count suggests. In practice, our TiME-m models achieve up to 1.6$\times$ speedup over XLM-R-Base and 5.5$\times$ over XLM-R-Large (see Table~\ref{table:performance_efficiency_summary}).
Detailed per-task results for all configurations are available in Appendix~\ref{sec:appendix}.

\begin{table*}[t]
\centering
\small
\setlength{\tabcolsep}{5pt}
\renewcommand{\arraystretch}{1.4}
\begin{tabular}{@{}l c c c c c c@{}}
\toprule
Model & Score & \shortstack{Latency \\(ms, BS=1)} & \shortstack{Peak TP \\(s/s) } & Peak Speedup
      & \shortstack{Opt. J/sample\\ (min. over BS)} & \shortstack{J/sample\\ Impr. ($\times$)} \\
\midrule
\multicolumn{7}{@{}l}{\textit{Baselines}} \\
HPLT (our eval)     & 84.27 & 11.35 & 606.7  & 1.6$\times$ & 0.6122 & 1.3$\times$ \\
XLM-R-Base          & 87.86 & 5.68  & 1168.5 & 3.2$\times$ & 0.2532 & 3.3$\times$ \\
mMiniLM-L6-H384     & 83.46 & \textbf{3.12} & \textbf{6229.4} & \textbf{16.9$\times$} & 0.0425 & 19.4$\times$ \\
XLM-R-Large         & \textbf{89.15} & 11.04 & 369.3  & 1.0$\times$ & 0.8260 & 1.0$\times$ \\
\cmidrule(l){1-7}
\multicolumn{7}{@{}l}{\textit{Our Students}} \\
\textbf{TiME-m (Ours)} & \textbf{87.71} & 2.85 & 1977.2 & 5.4$\times$ & 0.1249 & 6.6$\times$ \\
TiME-s (Ours)          & 85.16 & 2.86 & 5871.0 & 15.9$\times$ & 0.0443 & 18.7$\times$ \\
TiME-xs (Ours)         & 83.92 & \textbf{1.89} & \textbf{9321.1} & \textbf{25.2$\times$} & 0.0274 & 30.2$\times$ \\
*-hplt-m (Ours)        & 84.22 & 2.90 & 1515.6 & 4.1$\times$ & 0.1854 & 4.5$\times$ \\
\bottomrule
\end{tabular}
\caption{\textbf{Performance–efficiency trade-off.} Models are compared on their average performance score across all languages against key efficiency metrics. Latency is the averaged inference time at batch size 1. Peak Throughput is the average of the maximum achievable throughput for each model on each language, measured at its language-specific optimal batch size. We report the minimal (optimal) J/sample over batch size for each model and the improvement vs.\ XLM-R-Large.}
\label{table:performance_efficiency_summary}
\end{table*}

\subsection{Speed}
A key goal of our work is to produce models that are not only accurate but also fast enough for processing large amounts of data or for real-time applications. To visualize the performance-efficiency trade-off, Figure~\ref{fig:performance-efficiency-tradeoff-avg} presents the average NLP task scores against inference latency (at batch size 1) and throughput (at the optimal batch size for each model).

Our distilled \texttt{TiME-m} models, for instance, achieve an average score of 87.71 (vs. 87.86 for XLM-R-Base) with a latency of 2.9~ms (vs. 5.7~ms) and a throughput of 1799.8 sentences/s (vs. 1079.0 sentences/s). This positions them favorably on the efficiency frontier (approximated by the dashed line in Figure~\ref{fig:performance-efficiency-tradeoff-avg}), delivering performance comparable to \texttt{XLM-R-Base} but with substantially better efficiency. Our \texttt{TiME-m} models are also significantly faster than the original \texttt{XLM-R-Large} teacher. Our smaller student models, \texttt{TiME-en-s} and \texttt{en-hplt-xs}, offer further latency reductions and throughput improvements, making them suitable for scenarios where speed is the utmost priority. This demonstrates that our distillation pipeline is effective at creating models with improved performance-latency and performance-throughput trade-offs.

To situate our models within the landscape of practical NLP tooling, we compare them against spaCy’s transformer-based pipelines \cite{honnibal2020spacy} for supported languages. While a direct comparison is difficult, since a key contribution of our research is a distillation pipeline for languages that spaCy does not support off the shelf, Appendix \ref{ssec:spacy_comparison} provides detailed results for overlapping cases, showing that our models offer similar performance at significantly improved latency and throughput.

\subsection{Energy efficiency}
So far, we have looked at efficiency in terms of inference speed. Another effect of reducing model size can be improved energy efficiency, which we focus on in this section.

\begin{figure*}[t]
    \centering
    \includegraphics[width=0.99\linewidth]{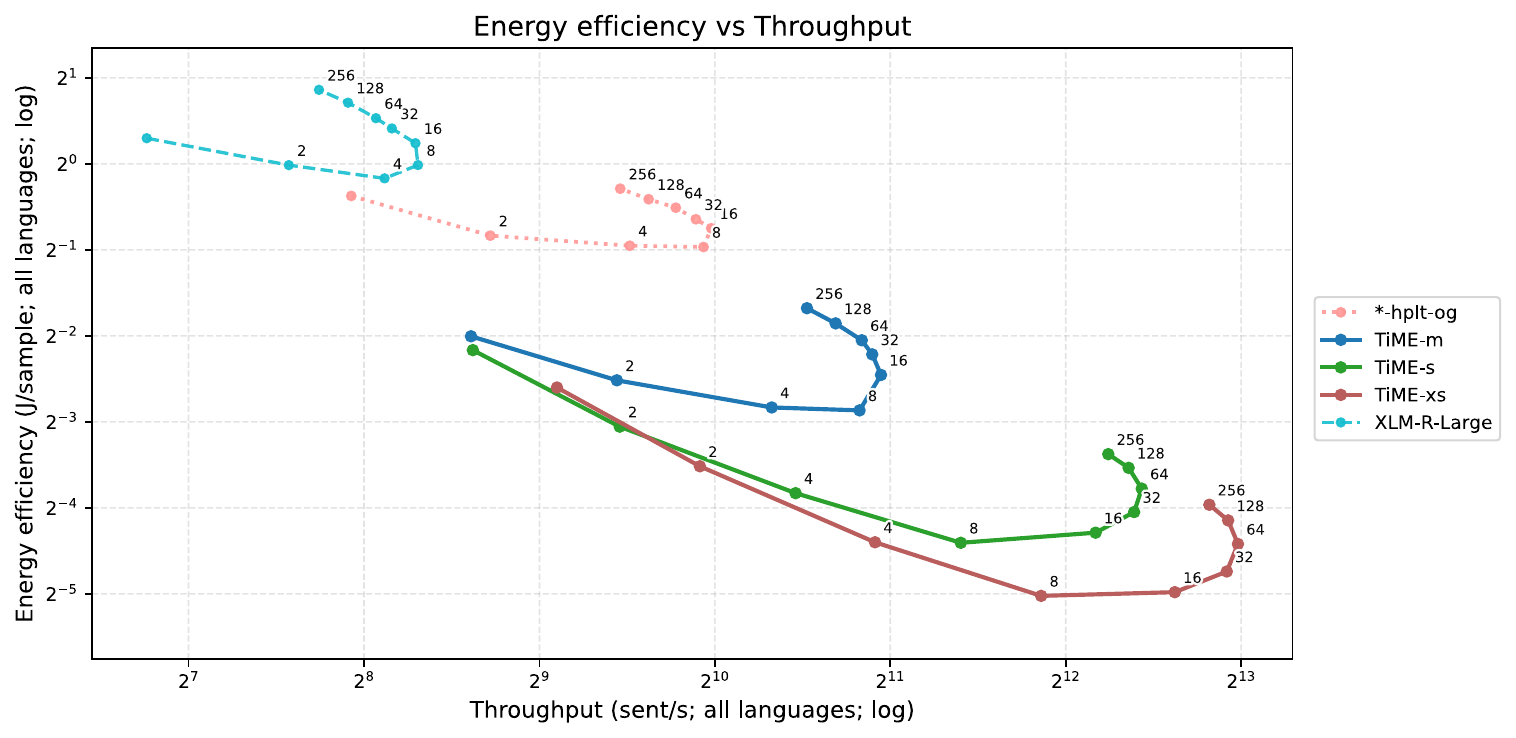}
    \caption{\textbf{Energy per sample vs.\ throughput, averaged across the seven core languages.}
    Each curve traces increasing batch sizes (markers annotated with the batch size).
    Lower is better on the y-axis (J/sample). The x-axis is logarithmic.}
    \label{fig:energy-vs-throughput-main}
\end{figure*}

\begin{figure*}[t]
    \centering
    \includegraphics[width=0.99\linewidth]{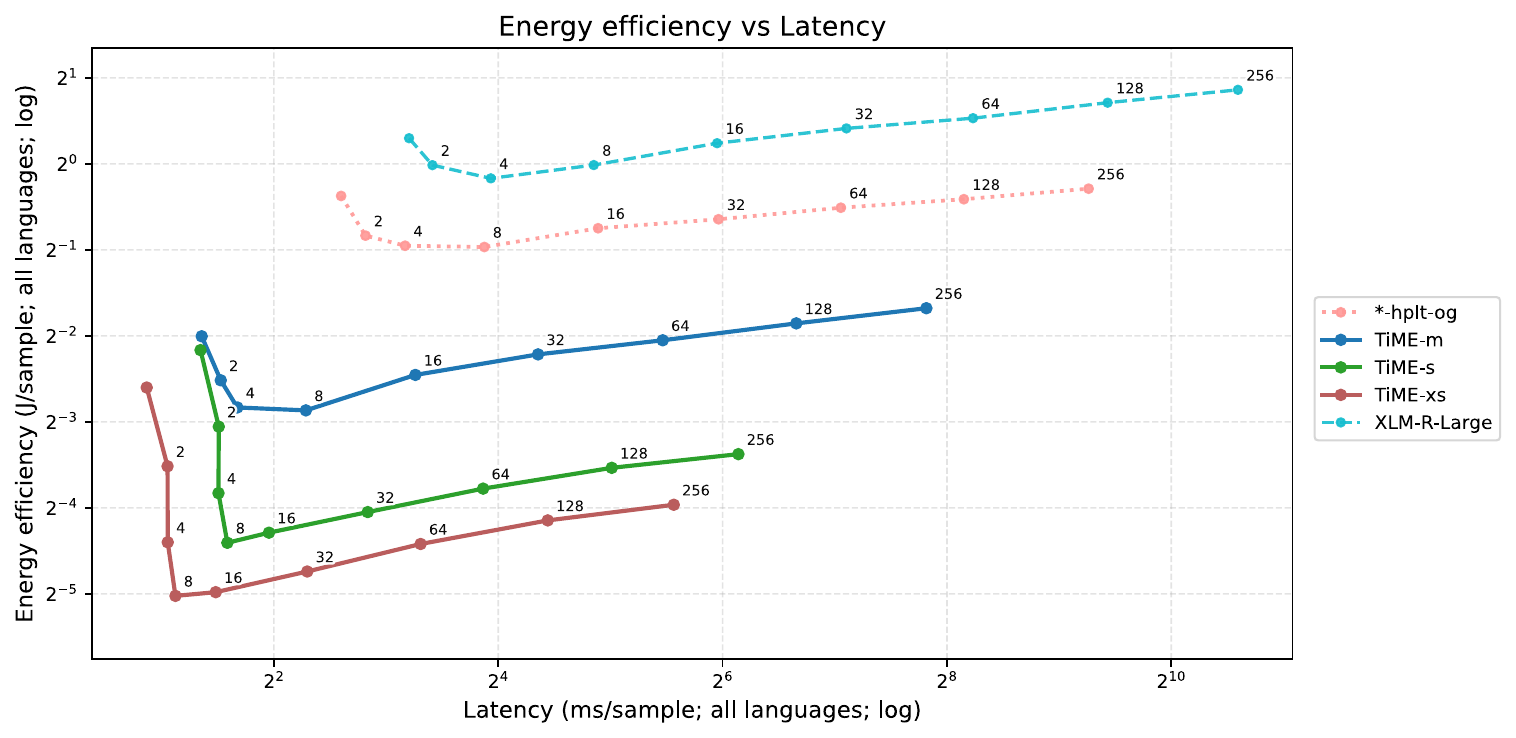}
    \caption{\textbf{Energy per sample vs.\ latency for different models and batch sizes.} The plot is log-log and averaged across the seven core languages.}
    \label{fig:energy-vs-latency-main}
\end{figure*}

\paragraph{Energy consumption and speed.}
Figures~\ref{fig:energy-vs-throughput-main} and \ref{fig:energy-vs-latency-main} show energy per sample (J/sample) against throughput and latency, respectively, at different batch sizes. We only include a selection of models in the plots for readability reasons; numbers for all models can be found in Tables~\ref{table:main_summary_condensed_full} and \ref{table:performance_efficiency_summary}. We see that the speed improvements of our \texttt{TiME} models translate into substantial energy efficiency improvements across batch sizes. Within a model, energy efficiency generally improves with throughput, though interestingly at the very highest throughput efficiency suffers slightly.

\begin{figure*}[t]
    \centering
    \includegraphics[width=0.78\linewidth]{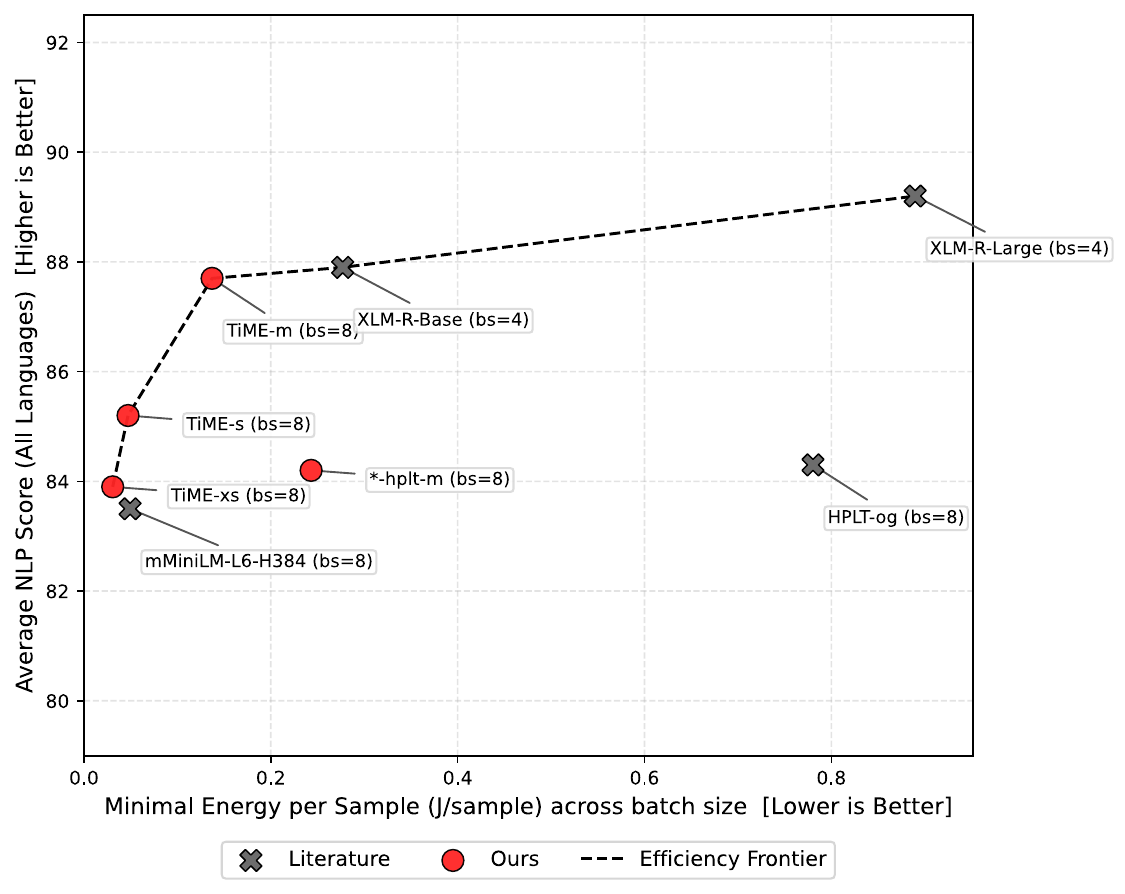}
    \caption{\textbf{NLP score vs.\ minimal energy per sample.} Energy consumption is measured for each model at the batch size that minimizes energy consumption per sample.}
    \label{fig:score-vs-energy}
\end{figure*}

\paragraph{Score--energy trade-off.}
Figure~\ref{fig:score-vs-energy} summarizes the accuracy–efficiency landscape when each model is operated at its most energy-efficient batch size.
The \texttt{TiME} models sit at the Pareto frontier, pairing low energy consumption per sample with strong task scores.

\subsection{Question Answering}
We evaluate the English and German models on the MLQA benchmark \cite{lewis-etal-2020-mlqa} to assess if the distillation preserves knowledge beyond core NLP tasks (the two languages that overlap with those covered by MLQA). The results in Table~\ref{tab:qa_results} show that the distilled models retain a large fraction of their teacher's performance. The \texttt{TiME-en-m} student, for instance, closes much of the performance gap to `XLM-R-Large`, indicating that the distillation process successfully transfers the ability to perform extractive question answering.

\begin{table}[h!]
\centering
\begin{tabular}{lccc}
\toprule
Language & Model & F1 & EM \\
\midrule
\multirow{3}{*}{English (en)} 
& TiME-en-xs & 70.00 & 56.22 \\ %
& TiME-en-s & 75.07 & 61.46 \\ %
& TiME-en-m & 81.15 & 67.69\\ %
& XLM-R-Base & 81.33 & 67.92 \\ %
& XLM-R-Large & 84.32 & 71.13 \\ %
\midrule
\multirow{3}{*}{German (de)} 
& TiME-de-xs & 50.34 & 33.74 \\ %
& TiME-de-s & 56.70 & 39.14 \\ %
& TiME-de-m & 61.05 & 42.28 \\ %
& XLM-R-Base & 59.95 & 41.95 \\ %
& XLM-R-Large & 65.81 & 45.87 \\ %
\bottomrule
\end{tabular}
\vspace{8pt}
\caption{\textbf{MLQA results for English and German.} Student models (`-s` and `-m`) are compared against the teacher.}
\label{tab:qa_results}
\end{table}

\section*{Limitations}

While our study demonstrates a practical pipeline for creating efficient Transformer-based encoders, we acknowledge several limitations.

\paragraph{Architectural Choices and Data.} Our selection of student model sizes (xs, s, m) was made to provide a practical set of options along the efficiency-performance curve. However, a more systematic architectural search was beyond the scope of this work and could yield further Pareto-optimal models. Similarly, we did not perform an ablation on the minimum amount of monolingual data required for effective distillation, which would be valuable for guiding future work on extremely low-resource languages.

\paragraph{NLP Score Differences Between Languages.} Our evaluation highlights that the performance-efficiency trade-off varies across languages. For instance, the performance drop for low-resource Irish (ga) and morphologically complex Hungarian (hu) was more pronounced in our smallest models, suggesting that certain linguistic properties might be more challenging to retain during compression. A deeper analysis of these trade-offs is a promising direction for future work.

\paragraph{Distillation Hyperparameters.} The number of relation heads ($A_r$) was set following the original MiniLMv2 work. A detailed ablation on this hyperparameter for each language could provide further optimization but was not performed in this study.

\section{Conclusion}

In this work, we have shown that monolingual distillation from multilingual teachers can lead to very efficient but still powerful models for common NLP tasks.
We have built a pipeline that is practical and effective across a wide range of languages, including those that are typically underserved. The resulting models are immediately useful in real-world scenarios, offering a drop-in, efficient alternative to legacy NLP pipelines.

Our models lie on the Pareto frontier of NLP taks performance on one hand and speed and energy efficiency on the other. Our extensive benchmarking allows practitioners to make informed decisions about which model to choose for their particular needs.


\bibliographystyle{plainnat}
\bibliography{custom}

\clearpage
\onecolumn
\appendix

\section{Additional Figures and Detailed Results}
\label{sec:appendix}
\begin{figure}
\centering\includegraphics[
  width=\linewidth,
  trim=0cm 0cm 6.0cm 0cm,
  clip
]{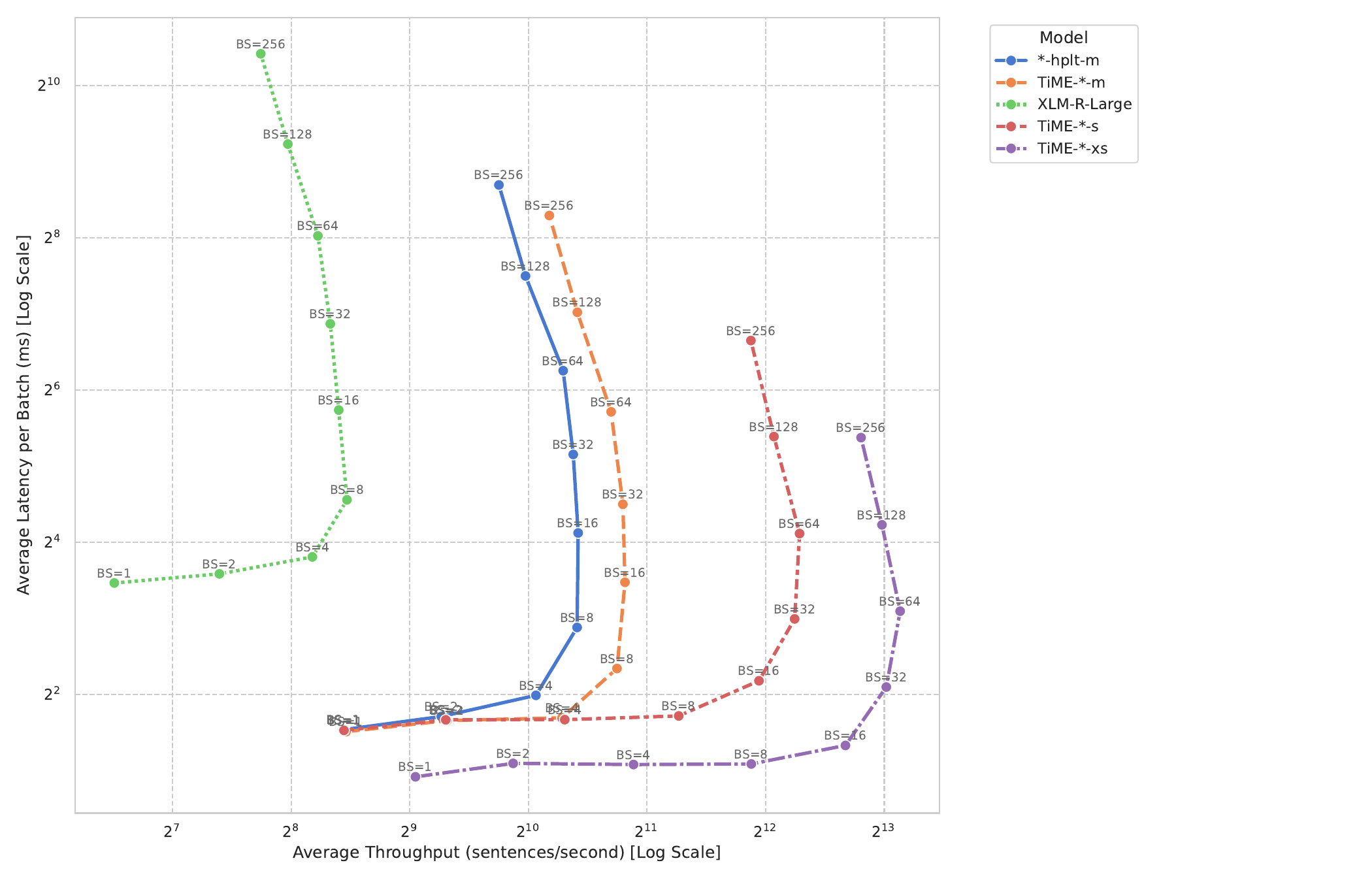}
\caption{\textbf{Latency vs. throughput for different batch sizes.} Language-specific values were obtained on the Wikipedia datasets for the 7 core languages.}
\label{fig:latency-vs-throughput-batchsize}
\end{figure}

\begin{figure*}[htbp]
    \centering
    \begin{subfigure}[b]{0.49\textwidth}
        \centering
        \includegraphics[width=\linewidth]{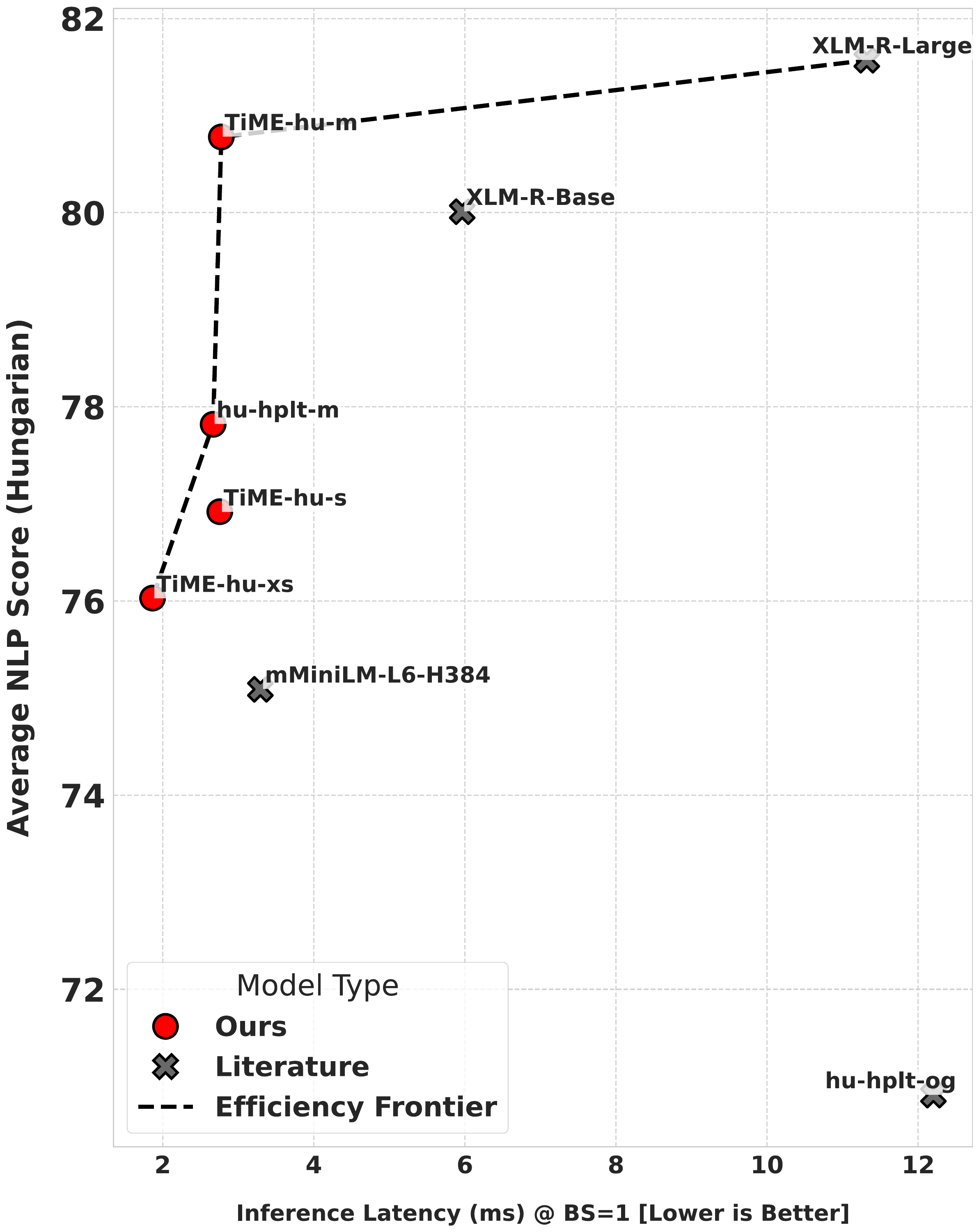}
        \caption{Performance vs. Latency}
        \label{fig:hu_latency_appendix}
    \end{subfigure}
    \hfill
    \begin{subfigure}[b]{0.49\textwidth}
        \centering
        \includegraphics[width=\linewidth]{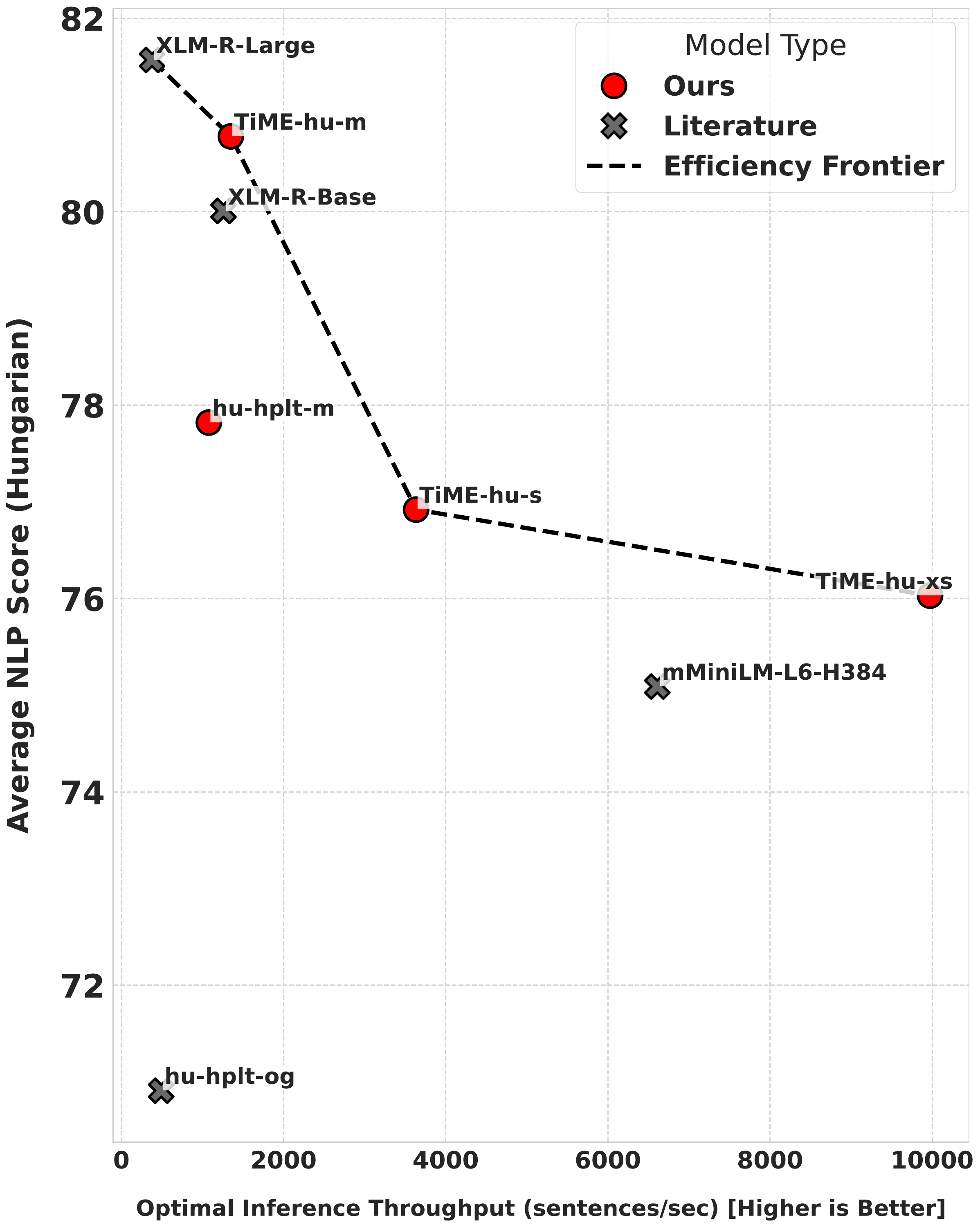}
        \caption{Performance vs. Throughput}
        \label{fig:hu_throughput_appendix}
    \end{subfigure}
    \caption{\textbf{Performance--efficiency trade-off for Hungarian models at batch size 1.} For the latency plot (a), the optimal position is the upper-left (high score, low latency). For the throughput plot (b), the optimal position is the upper-right (high score, high throughput).}
    \label{fig:hu_tradeoff_appendix}
\end{figure*}

\begin{figure*}[htbp]
    \centering
    \begin{subfigure}[b]{0.49\textwidth}
        \centering
        \includegraphics[width=\linewidth]{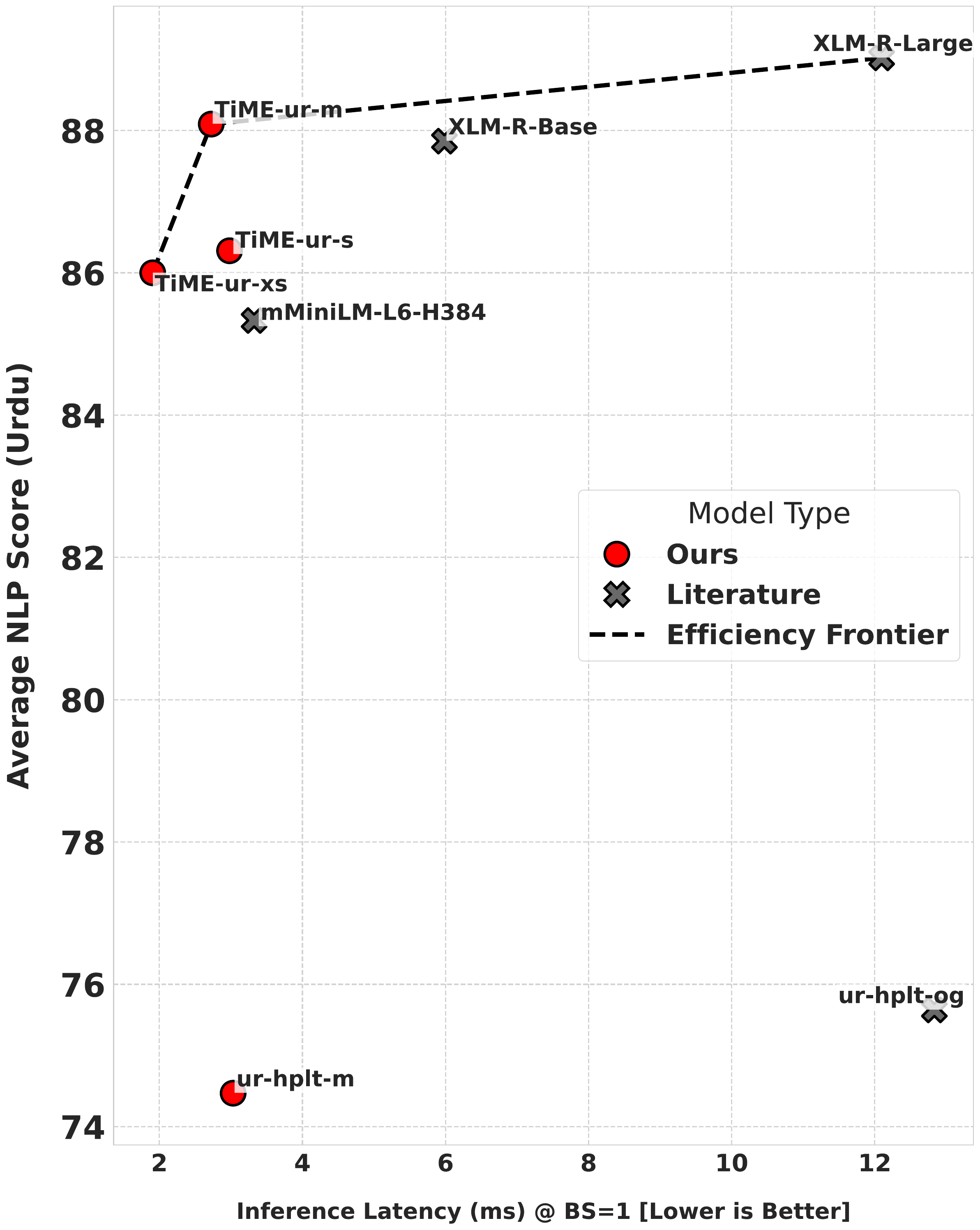}
        \caption{Performance vs. Latency}
        \label{fig:ur_latency_appendix}
    \end{subfigure}
    \hfill
    \begin{subfigure}[b]{0.49\textwidth}
        \centering
        \includegraphics[width=\linewidth]{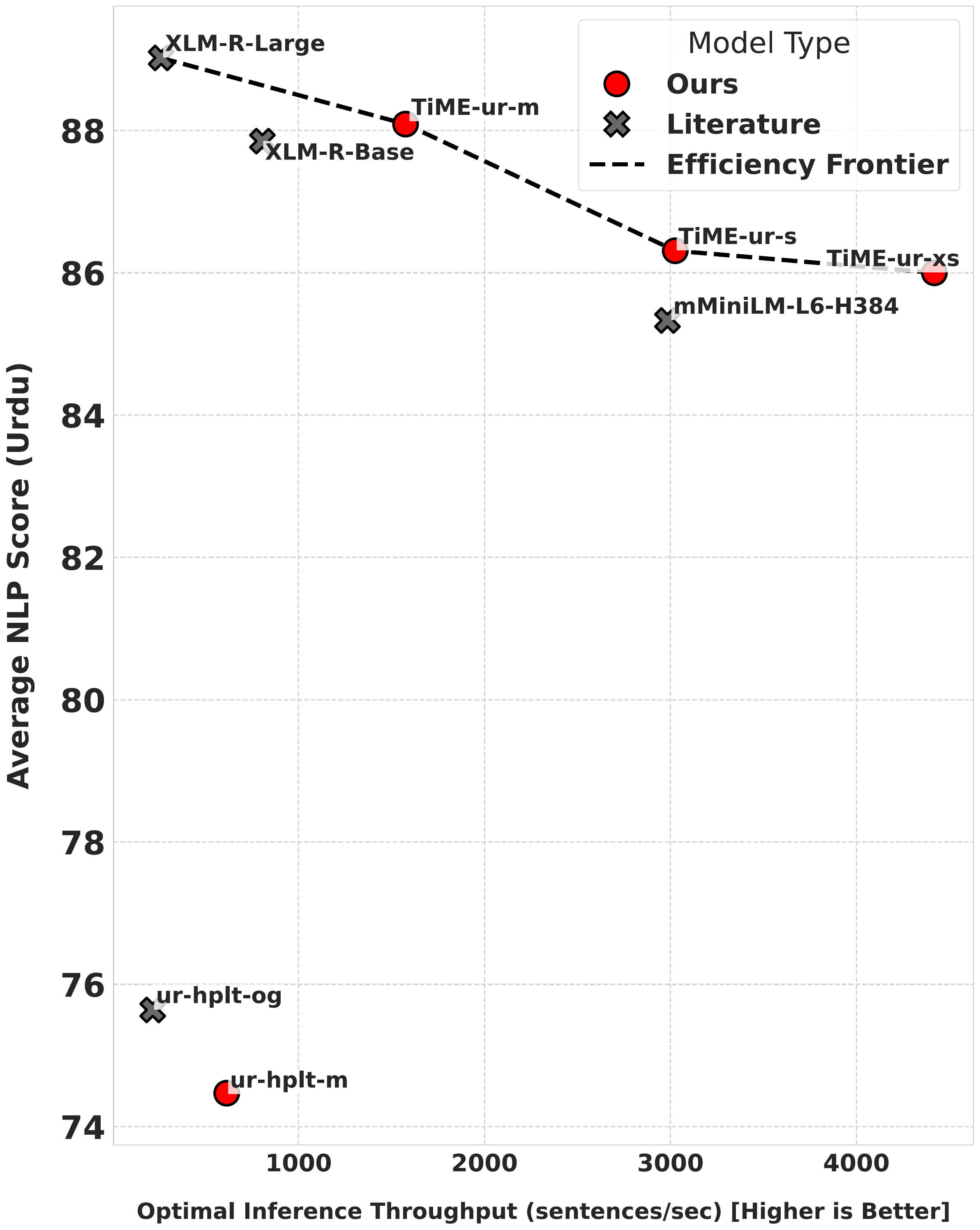}
        \caption{Performance vs. Throughput}
        \label{fig:ur_throughput_appendix}
    \end{subfigure}
    \caption{\textbf{Performance--efficiency trade-off for Urdu models at batch size 1.} For the latency plot (a), the optimal position is the upper-left (high score, low latency). For the throughput plot (b), the optimal position is the upper-right (high score, high throughput).}
    \label{fig:ur_tradeoff_appendix}
\end{figure*}

\begin{figure*}[htbp]
    \centering
    \begin{subfigure}[b]{0.49\textwidth}
        \centering
        \includegraphics[width=\linewidth]{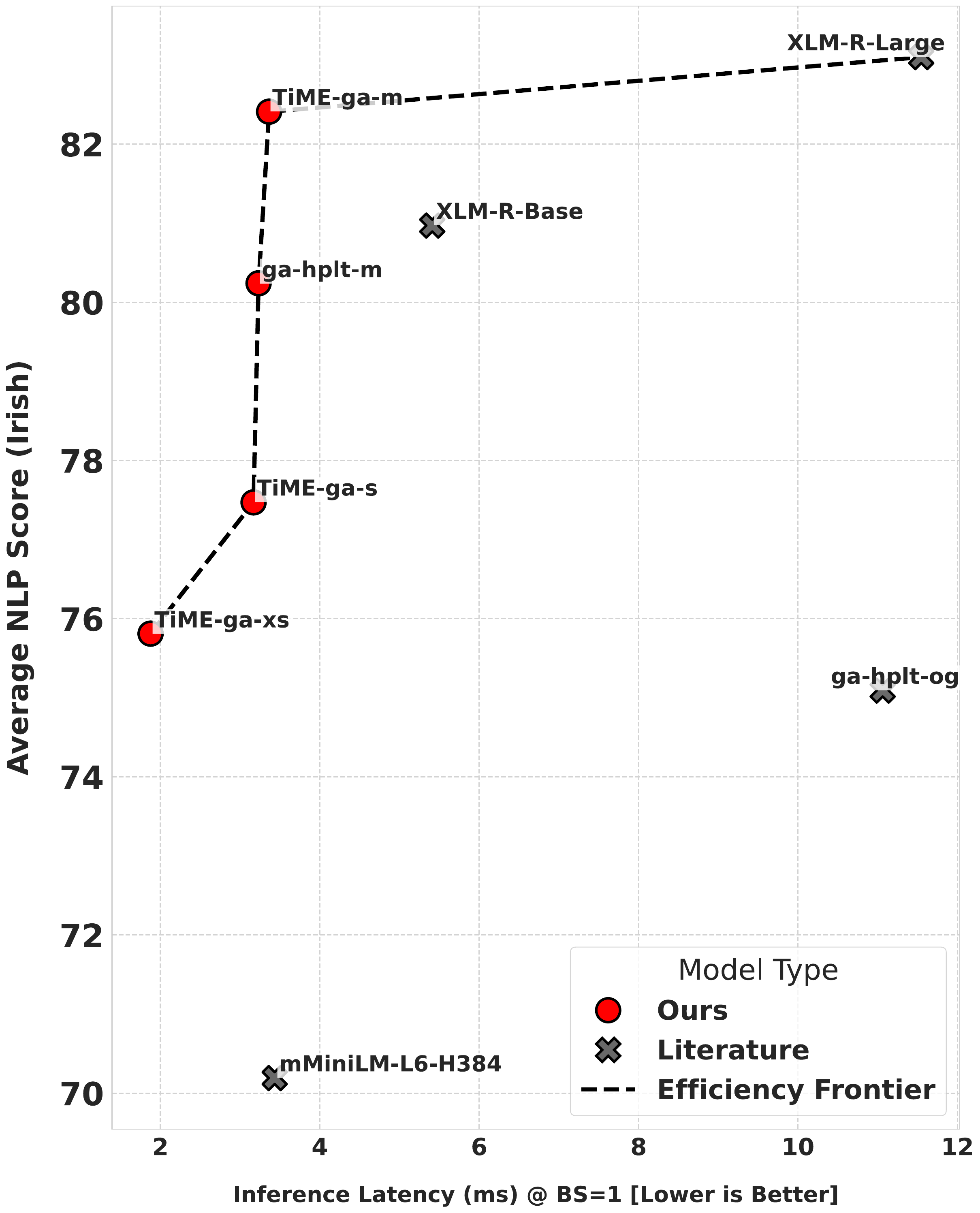}
        \caption{Performance vs. Latency}
        \label{fig:ga_latency_appendix}
    \end{subfigure}
    \hfill
    \begin{subfigure}[b]{0.49\textwidth}
        \centering
        \includegraphics[width=\linewidth]{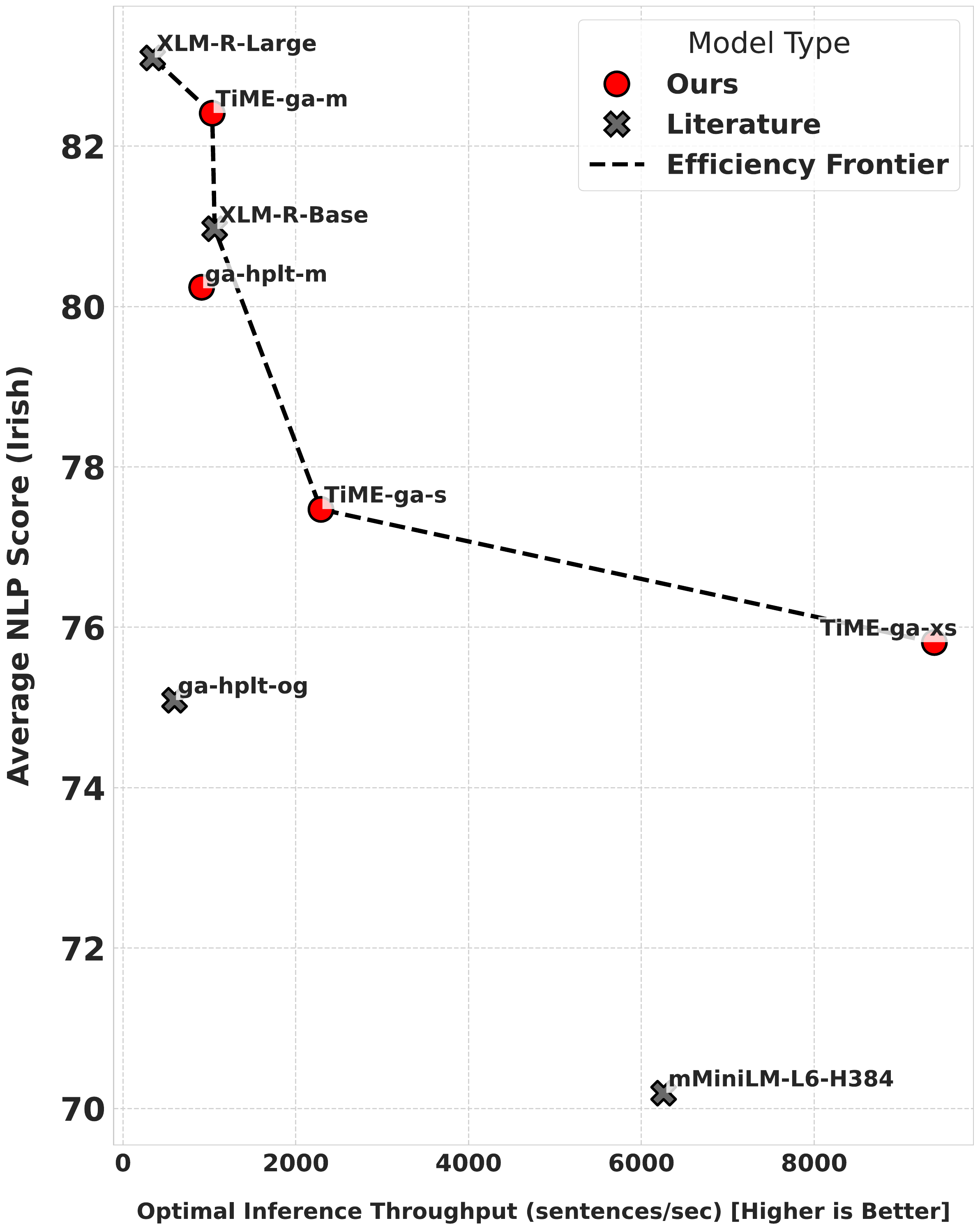}
        \caption{Performance vs. Throughput}
        \label{fig:ga_throughput_appendix}
    \end{subfigure}
    \caption{\textbf{Performance--efficiency trade-off for Irish models at batch size 1.} For the latency plot (a), the optimal position is the upper-left (high score, low latency). For the throughput plot (b), the optimal position is the upper-right (high score, high throughput).}
    \label{fig:ga_tradeoff_appendix}
\end{figure*}

In Table~\ref{table:appendix_detailed_scores_time} we show the detailed per-task and per-language performance on the NLP tasks. Note that for English we also compare with en-MiniLM-L6-H768 \cite{wang2021minilmv2multiheadselfattentionrelation}, a model distilled from the monolingual RoBERTa-Large. Figures \ref{fig:en_tradeoff}--\ref{fig:ga_tradeoff_appendix} are per-language versions of Figure \ref{fig:performance-efficiency-tradeoff-avg}, showing the trade-offs between NLP score on one hand, and latency and throughput on the other hand.
Figure~\ref{fig:latency-vs-throughput-batchsize} shows the relationship between batch size and throughput.

\subsection{Comparison with spaCy}
\label{ssec:spacy_comparison}

To situate our models within the landscape of production-ready tools, we benchmark them against spaCy \citep{honnibal2020spacy}, a widely-adopted industry standard for efficient NLP. It is important to note that SpaCy's transformer pipelines (\texttt{\_trf}) are not novel architectures but provide a consistent, production-optimized API for fine-tuning established pre-trained models. For the languages we evaluate, the underlying models are well-known encoders: the English pipeline (\texttt{en\_core\_web\_trf}) is based on RoBERTa \citep{liu2019roberta}; the German pipeline (\texttt{de\_dep\_news\_trf}) uses a cased German BERT \citep{bertbasegermancased}, itself based on the original BERT architecture \citep{devlin2019bert}; the French pipeline (\texttt{fr\_dep\_news\_trf}) leverages CamemBERT \citep{martin2019camembert}; and the Danish pipeline (\texttt{da\_core\_news\_trf}) is built upon DanskBERT \citep{snaebjarnarson-etal-2023-transfer}. We compare our \texttt{TiME-m}, \texttt{TiME-s}, and \texttt{TiME-xs} models against these pipelines.

\begin{figure*}[htbp]
    \centering
    \begin{subfigure}[b]{0.49\textwidth}
        \centering
        \includegraphics[width=\linewidth]{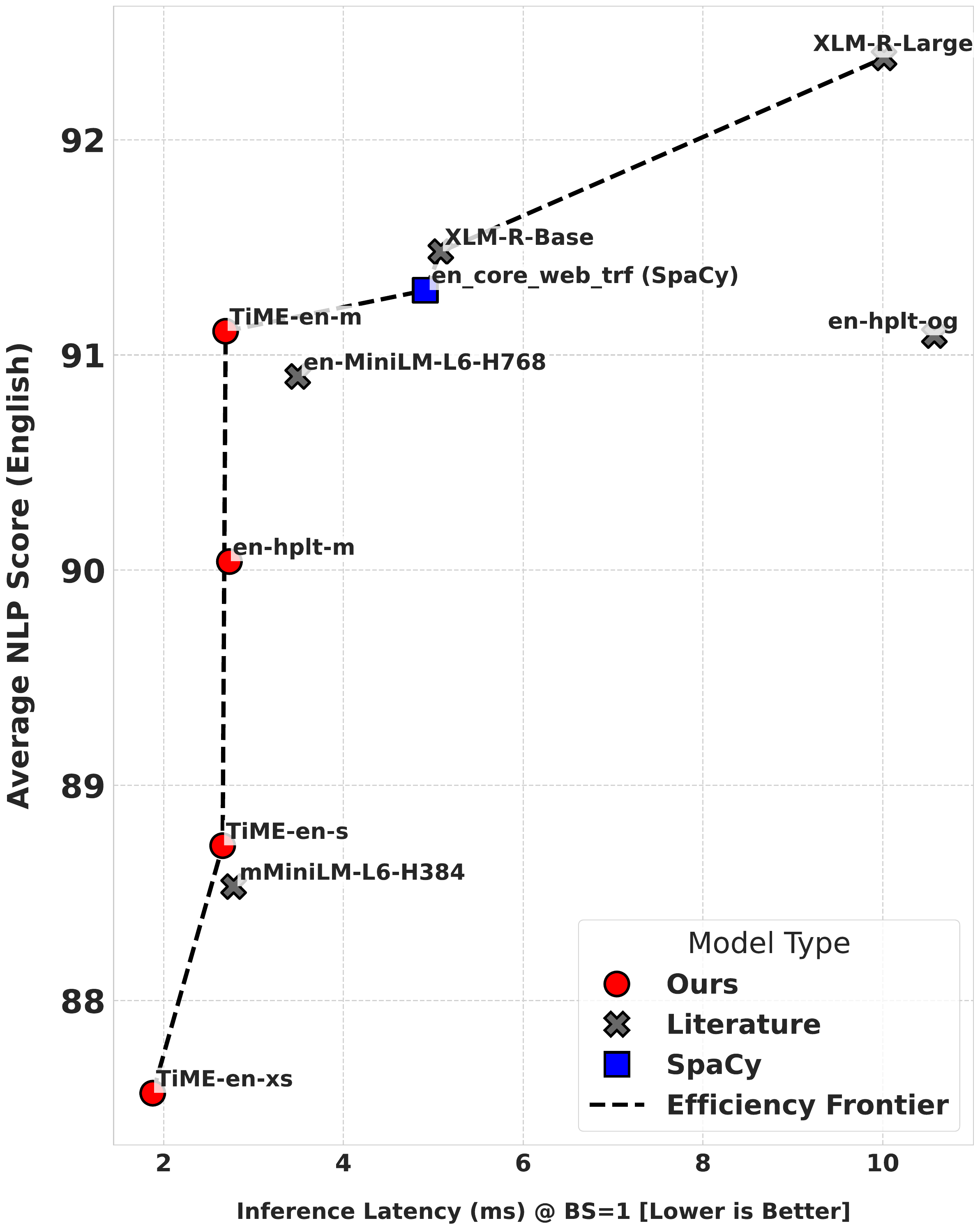}
        \caption{Performance vs. Latency}
        \label{fig:en_latency_appendix}
    \end{subfigure}
    \hfill 
    \begin{subfigure}[b]{0.49\textwidth}
        \centering
        \includegraphics[width=\linewidth]{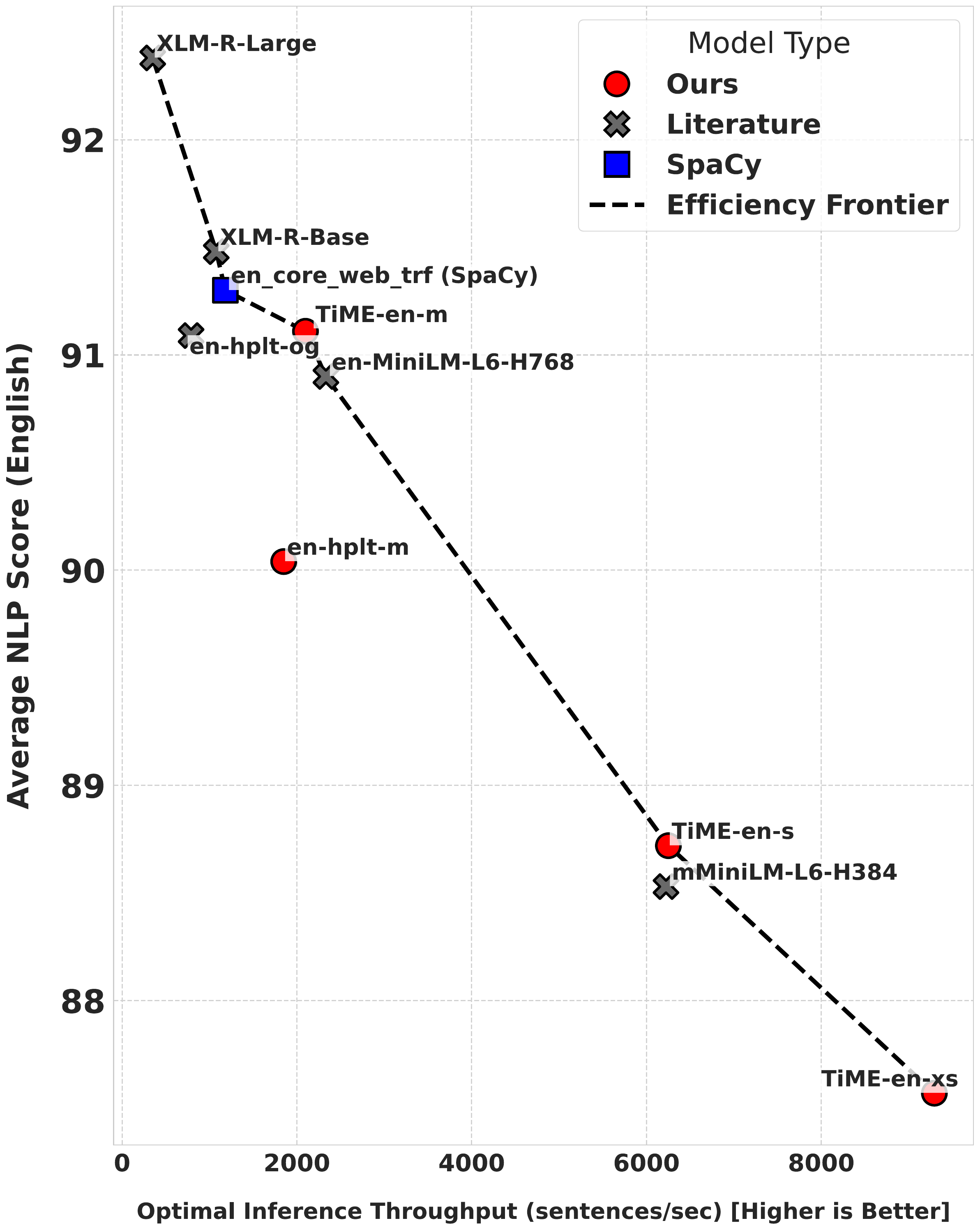}
        \caption{Performance vs. Throughput}
        \label{fig:en_throughput_appendix}
    \end{subfigure}
    \caption{\textbf{Performance--efficiency trade-off for English models at batch size 1.} For the latency plot (a), the optimal position is the upper-left (high score, low latency). For the throughput plot (b), the optimal position is the upper-right (high score, high throughput).}
    \label{fig:en_tradeoff}
\end{figure*}

\begin{figure*}[htbp]
    \centering
    \begin{subfigure}[b]{0.49\textwidth}
        \centering
        \includegraphics[width=\linewidth]{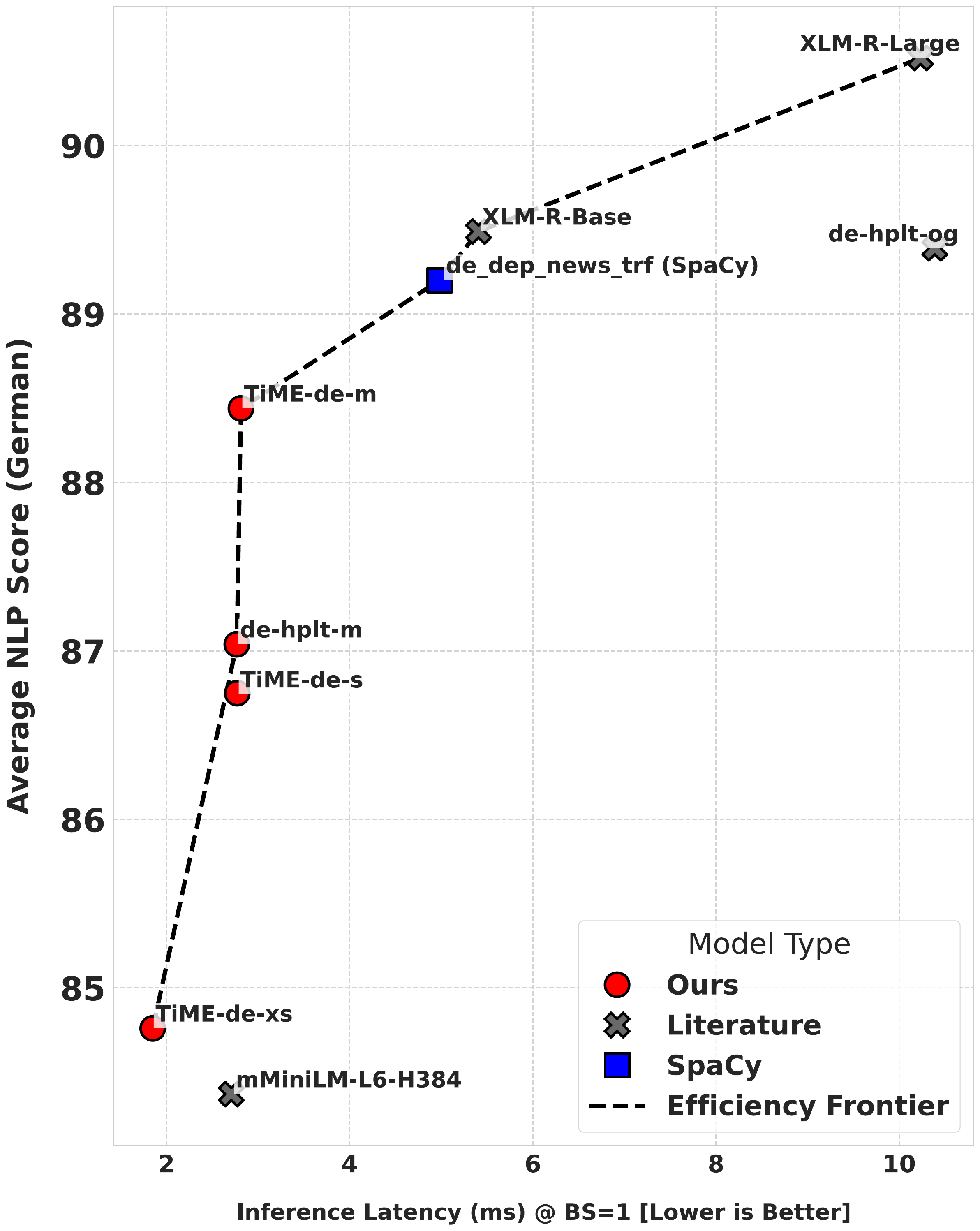}
        \caption{Performance vs. Latency}
        \label{fig:de_latency_appendix}
    \end{subfigure}
    \hfill
    \begin{subfigure}[b]{0.49\textwidth}
        \centering
        \includegraphics[width=\linewidth]{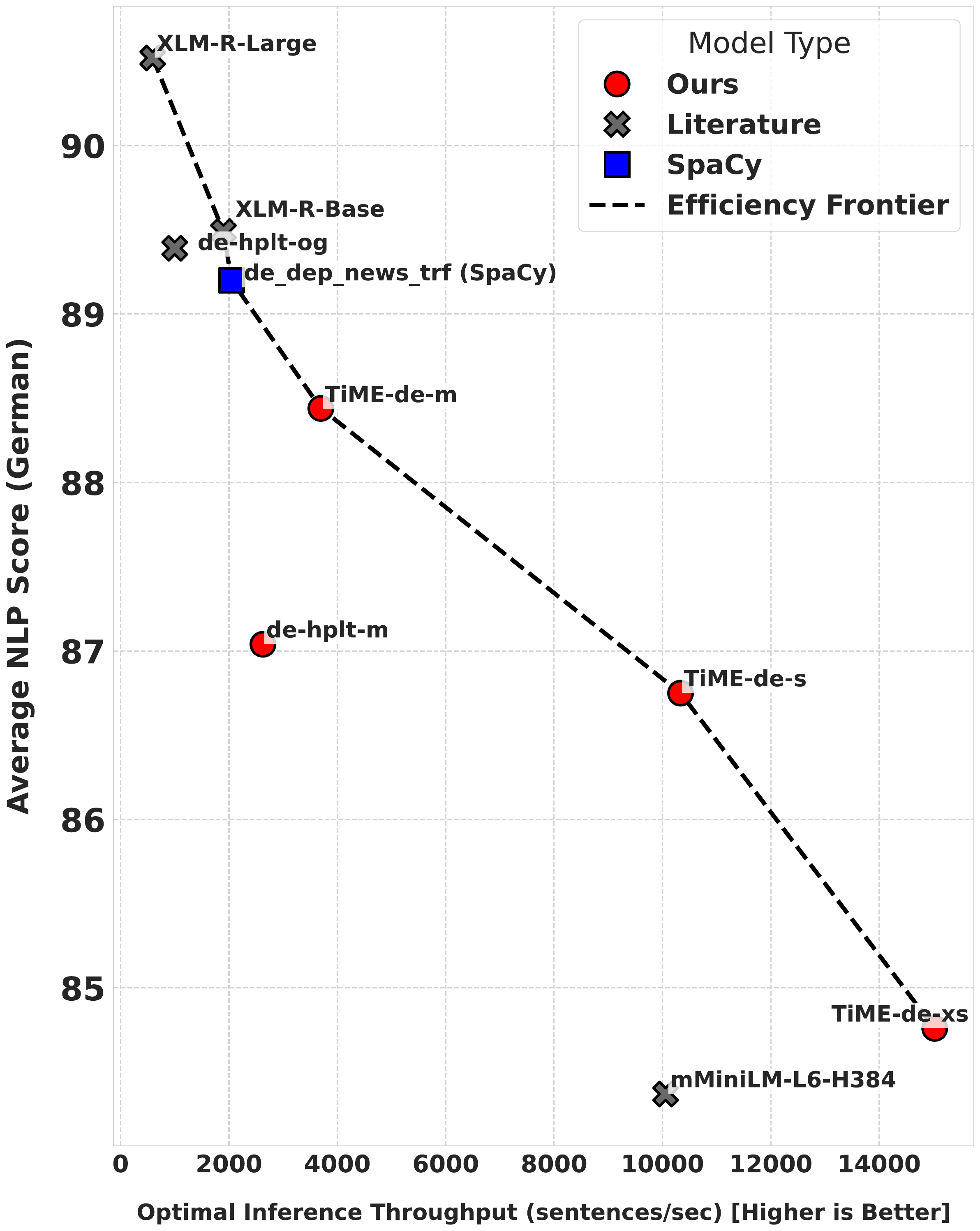}
        \caption{Performance vs. Throughput}
        \label{fig:de_throughput_appendix}
    \end{subfigure}
    \caption{\textbf{Performance--efficiency trade-off for German models at batch size 1.} For the latency plot (a), the optimal position is the upper-left (high score, low latency). For the throughput plot (b), the optimal position is the upper-right (high score, high throughput).}
    \label{fig:de_tradeoff_appendix}
\end{figure*}

\begin{figure*}[htbp]
    \centering
    \begin{subfigure}[b]{0.49\textwidth}
        \centering
        \includegraphics[width=\linewidth]{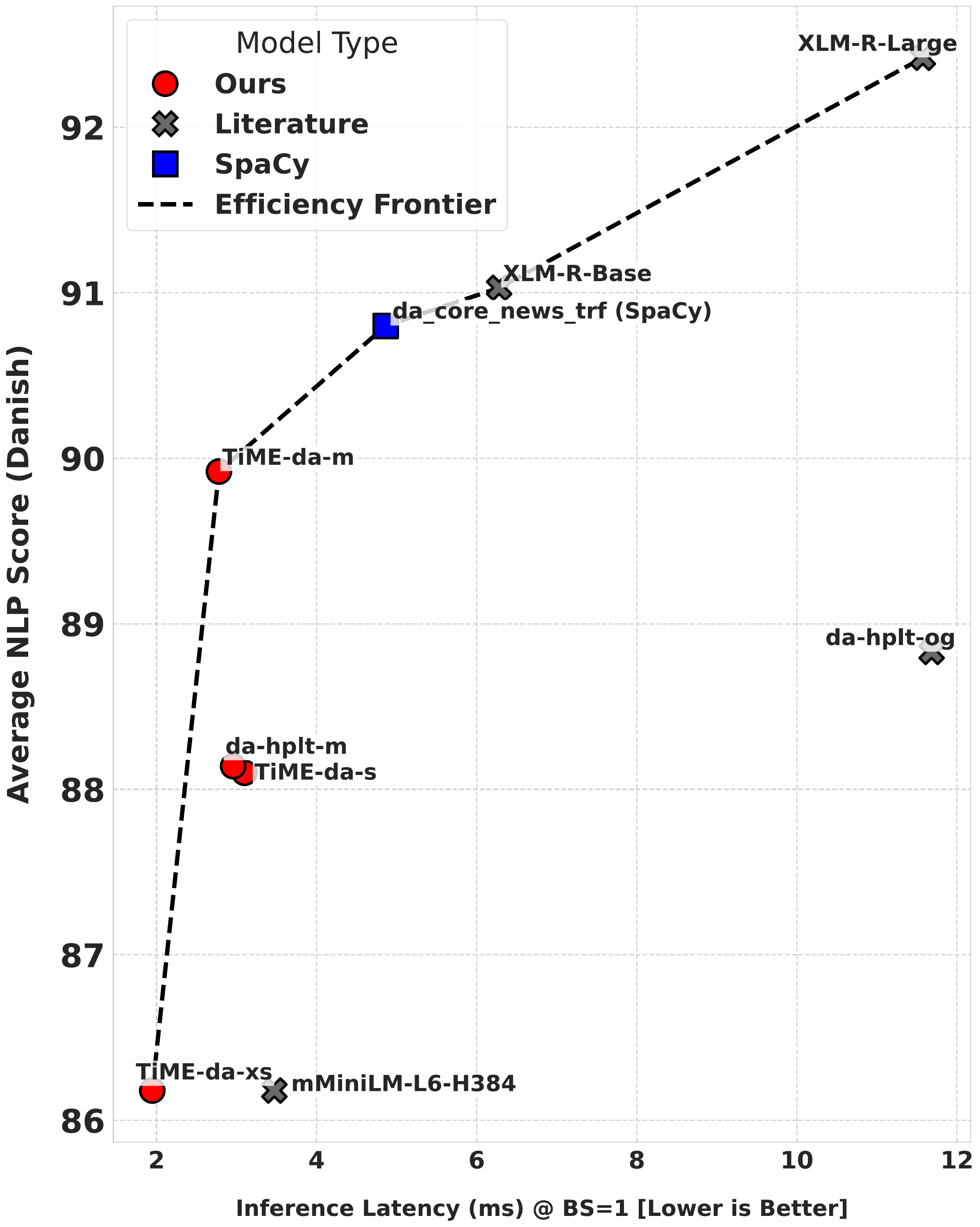}
        \caption{Performance vs. Latency}
        \label{fig:da_latency_appendix}
    \end{subfigure}
    \hfill
    \begin{subfigure}[b]{0.49\textwidth}
        \centering
        \includegraphics[width=\linewidth]{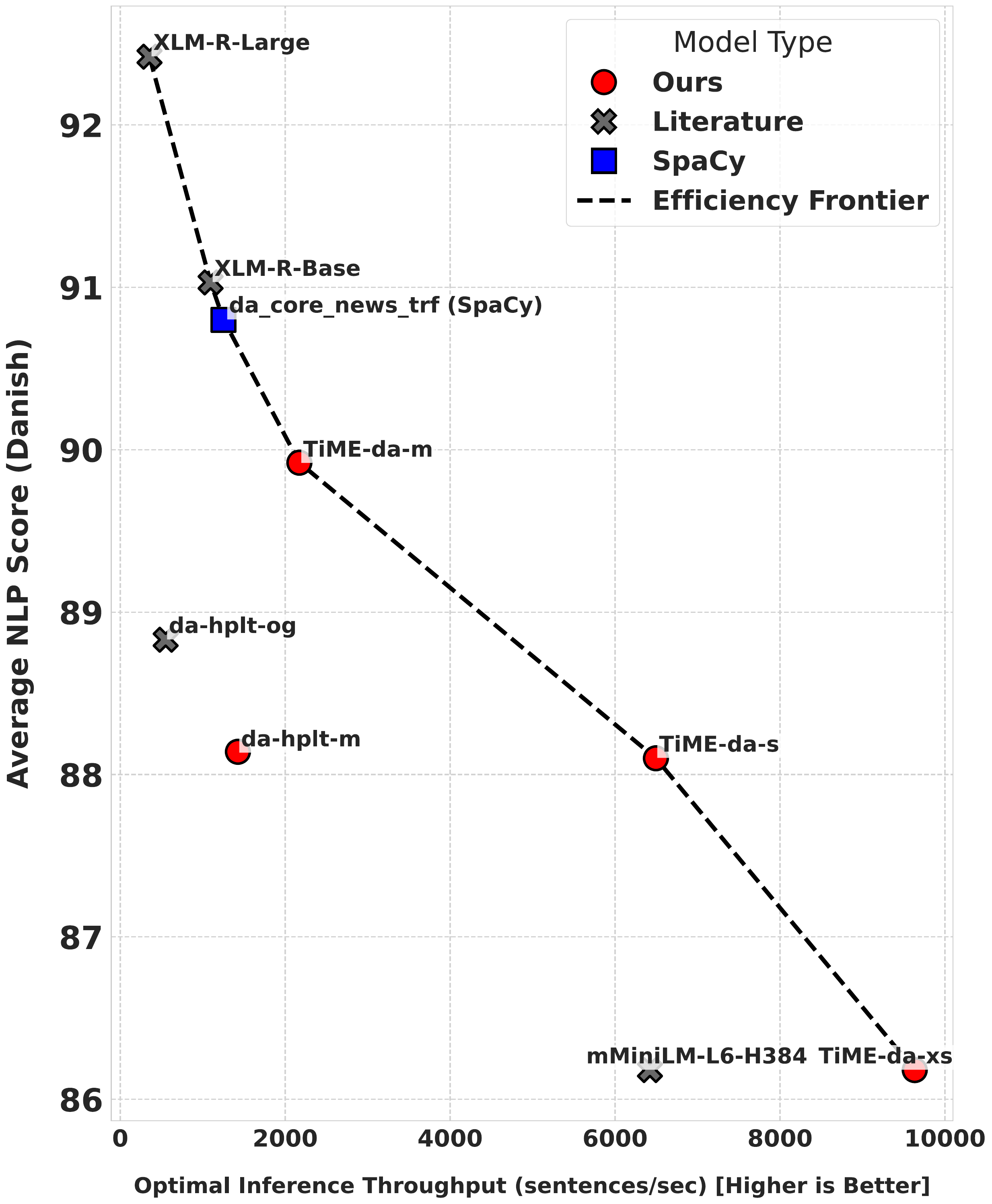}
        \caption{Performance vs. Throughput}
        \label{fig:da_throughput_appendix}
    \end{subfigure}
    \caption{\textbf{Performance--efficiency trade-off for Danish models at batch size 1.} For the latency plot (a), the optimal position is the upper-left (high score, low latency). For the throughput plot (b), the optimal position is the upper-right (high score, high throughput).}
    \label{fig:da_tradeoff_appendix}
\end{figure*}

\begin{figure*}[htbp]
    \centering
    \begin{subfigure}[b]{0.49\textwidth}
        \centering
        \includegraphics[width=\linewidth]{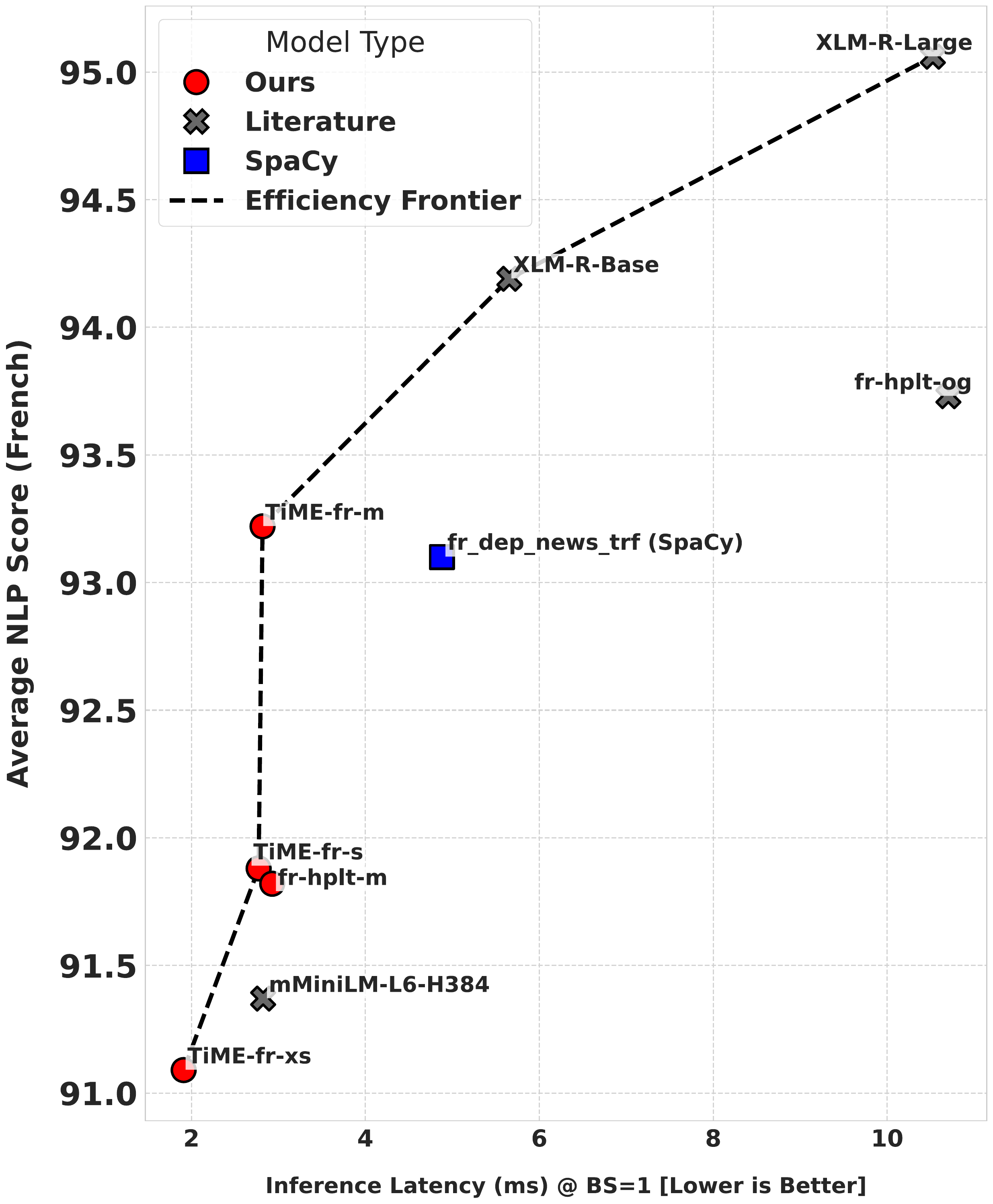}
        \caption{Performance vs. Latency}
        \label{fig:fr_latency_appendix}
    \end{subfigure}
    \hfill
    \begin{subfigure}[b]{0.49\textwidth}
        \centering
        \includegraphics[width=\linewidth]{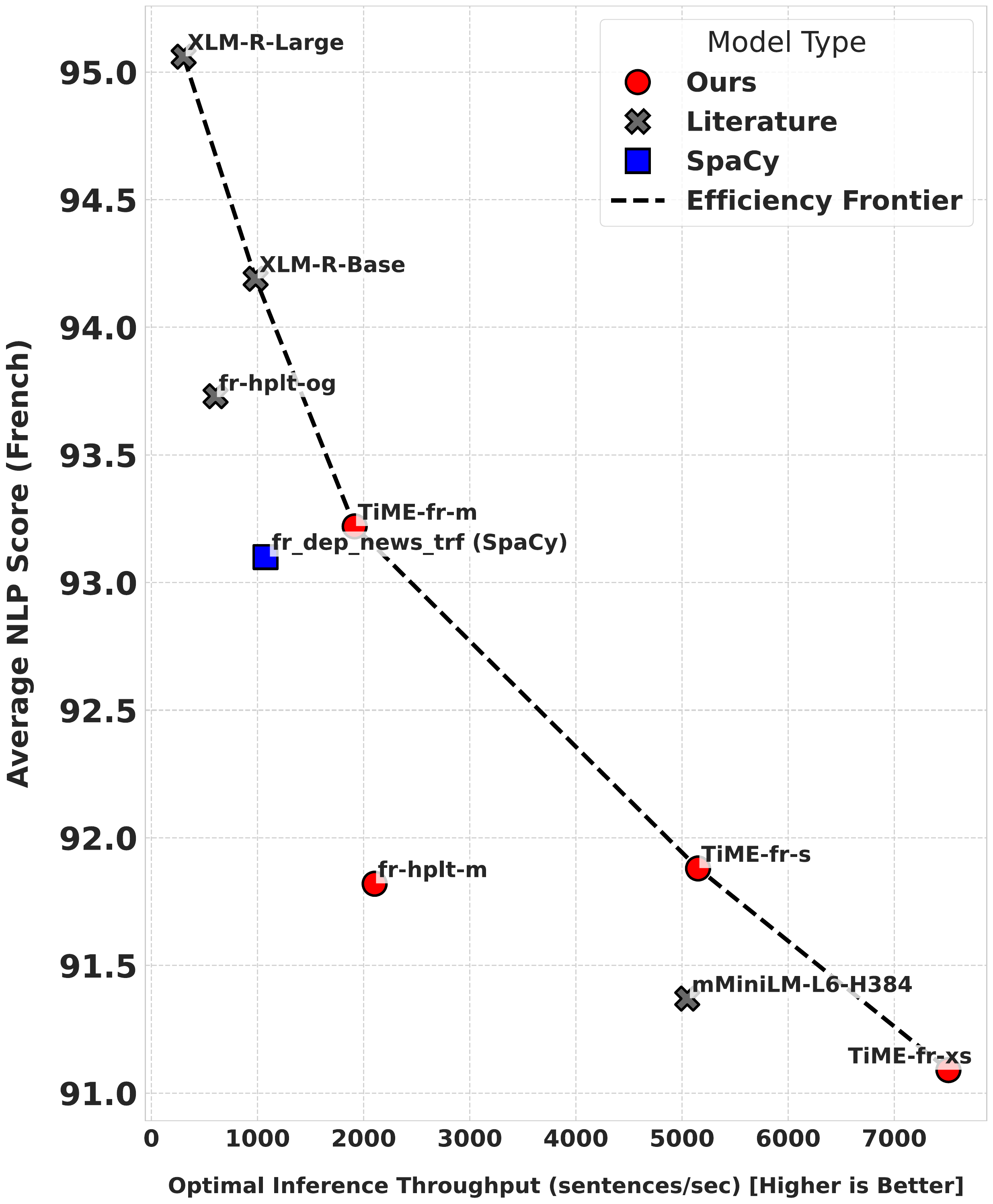}
        \caption{Performance vs. Throughput}
        \label{fig:fr_throughput_appendix}
    \end{subfigure}
    \caption{\textbf{Performance--efficiency trade-off for French models at batch size 1.} For the latency plot (a), the optimal position is the upper-left (high score, low latency). For the throughput plot (b), the optimal position is the upper-right (high score, high throughput).}
    \label{fig:fr_tradeoff_appendix}
\end{figure*}

The results are presented in Table~\ref{table:spacy_comparison}, with corresponding plots in Figures \ref{fig:en_tradeoff}, \ref{fig:de_tradeoff_appendix}, \ref{fig:da_tradeoff_appendix} and \ref{fig:fr_tradeoff_appendix}. In terms of accuracy, our medium-sized \texttt{TiME-m} models are highly competitive, achieving scores that are close to spaCy's \texttt{\_trf} models. This demonstrates that our distillation pipeline can produce models that match the quality of state-of-the-art production systems.

The primary advantage of our distilled models becomes evident in the efficiency metrics. For real-time applications, our models demonstrate significantly lower latency. Our \texttt{TiME-xs} model has less than half the latency of its spaCy counterpart (e.g., 1.9\,ms vs.\ 5.1\,ms for English). Throughput gains at each model’s optimal batch size are substantial: \texttt{TiME-en-xs} reaches 6361.6 vs.\ 1330.0 s/s on English (4.78$\times$), \texttt{TiME-de-xs} 10651.7 vs.\ 2046.3 on German (5.21$\times$), \texttt{TiME-fr-xs} 5077.3 vs.\ 1270.5 on French (4.00$\times$), and \texttt{TiME-da-xs} 5794.3 vs.\ 1494.5 on Danish (3.88$\times$). This makes our models well-suited for large-scale batch processing where computational cost and processing time matter.

\begin{table*}[!htbp]
\centering
\small
\setlength{\tabcolsep}{5pt} 
\renewcommand{\arraystretch}{1.1}
\begin{tabular}{@{}llccc@{}}
\toprule
Language & Model & Avg Score & Latency (ms) & Throughput (s/s) \\
\midrule
\multirow{4}{*}{English} 
& \textbf{TiME-en-m (ours)}  & 91.11 & 2.9 & 1539.3 \\
& \textbf{TiME-en-s (ours)}  & 88.72 & 2.9 & 4427.1 \\
& \textbf{TiME-en-xs (ours)} & 87.57 & 1.9 & 6361.6 \\
& en\_core\_web\_trf (spaCy) & 91.30 & 5.1 & 1330.0 \\
\midrule
\multirow{4}{*}{German} 
& \textbf{TiME-de-m (ours)}  & 88.44 & 2.9 & 3471.2 \\
& \textbf{TiME-de-s (ours)}  & 86.75 & 2.9 & 7545.8 \\
& \textbf{TiME-de-xs (ours)} & 84.76 & 1.9 & 10651.7 \\
& de\_dep\_news\_trf (spaCy) & 89.20 & 4.9 & 2046.3 \\
\midrule
\multirow{4}{*}{French} 
& \textbf{TiME-fr-m (ours)}  & 93.22 & 2.9 & 1454.4 \\
& \textbf{TiME-fr-s (ours)}  & 91.88 & 2.9 & 3589.8 \\
& \textbf{TiME-fr-xs (ours)} & 91.09 & 1.9 & 5077.3 \\
& fr\_dep\_news\_trf (spaCy) & 93.10 & 5.1 & 1270.5 \\
\midrule
\multirow{4}{*}{Danish} 
& \textbf{TiME-da-m (ours)}  & 89.92 & 2.9 & 1546.8 \\
& \textbf{TiME-da-s (ours)}  & 88.10 & 2.9 & 4264.2 \\
& \textbf{TiME-da-xs (ours)} & 86.18 & 1.9 & 5794.3 \\
& da\_core\_news\_trf (spaCy)& 90.80 & 5.6 & 1494.5 \\
\bottomrule
\end{tabular}
\caption{Comparison with spaCy pipelines for English, German, French, and Danish. 'Avg Score' is the average performance across four NLP tasks. Latency is measured in ms at batch size 1, and Throughput is the value at the optimal batch size for each model in sentences/sec.}
\label{table:spacy_comparison}
\end{table*}

\begin{longtable}{@{}llccccc@{}}
\caption{\textbf{Detailed NLP task performance for all 16 languages.} All scores on a 0--100 scale.}
\label{table:appendix_detailed_scores_time}\\
\toprule
Language & Model ID & NER & AllTags & Lemma & LAS & Avg Score \\
\midrule
\endfirsthead
\multicolumn{7}{c}{{\bfseries \tablename\ \thetable{} -- continued from previous page}}\\
\toprule
Language & Model ID & NER & AllTags & Lemma & LAS & Avg Score \\
\midrule
\endhead
\midrule
\multicolumn{7}{r@{}}{\textit{Continued on next page}}\\
\endfoot
\bottomrule
\endlastfoot

\multirow{7}{*}{\rotatebox[origin=c]{90}{Arabic (ar)}}
& TiME-xs           & 83.9 & 87.95 & 75.36 & 77.39 & 81.15 \\
& TiME-s            & 86.1 & 89.11 & 76.25 & 79.05 & 82.63 \\
& TiME-m            & 87.2 & 91.71 & 84.35 & 82.33 & 86.40 \\
\cmidrule(lr){2-7}
& XLM-R-Large       & 88.4 & 94.28 & 87.99 & 84.03 & 88.67 \\
& XLM-R-Base        & 86.7 & 92.60 & 84.46 & 82.44 & 86.55 \\
& mMiniLM-L6-H384   & 84.4 & 86.51 & 73.76 & 77.09 & 80.44 \\
& mMiniLM-L12-H384  & 85.8 & 88.52 & 74.84 & 79.01 & 82.04 \\
\midrule

\multirow{11}{*}{\rotatebox[origin=c]{90}{Danish (da)}}
& TiME-xs                 & 87.7 & 91.52 & 91.46 & 74.04 & 86.18 \\
& TiME-s                  & 89.2 & 92.80 & 92.32 & 76.11 & 87.61 \\
& TiME-m                  & 90.7 & 95.51 & 93.76 & 81.01 & 90.25 \\
& da-hplt-xs              & 83.6 & 92.77 & 94.10 & 59.29 & 82.44 \\
& da-hplt-m               & 88.7 & 95.71 & 95.92 & 72.24 & 88.14 \\
\cmidrule(lr){2-7}
& XLM-R-Large             & 93.2 & 96.91 & 95.63 & 84.37 & 92.53 \\
& XLM-R-Base              & 90.9 & 95.82 & 94.85 & 82.61 & 91.05 \\
& mMiniLM-L6-H384         & 88.3 & 90.11 & 91.62 & 74.70 & 86.18 \\
& mMiniLM-L12-H384        & 89.9 & 91.53 & 92.01 & 77.49 & 87.73 \\
& da-hplt-og              & 92.1 & 95.15 & 92.66 & 75.41 & 88.83 \\
\midrule
\clearpage
\multirow{11}{*}{\rotatebox[origin=c]{90}{German (de)}}
& TiME-xs                 & 82.9 & 80.80 & 94.50 & 80.82 & 84.76 \\
& TiME-s                  & 85.8 & 83.26 & 94.87 & 82.97 & 86.73 \\
& TiME-m                  & 86.9 & 86.79 & 96.10 & 84.79 & 88.65 \\
& de-hplt-xs              & 80.1 & 82.42 & 94.82 & 75.18 & 83.13 \\
& de-hplt-m               & 84.1 & 86.88 & 96.11 & 81.05 & 87.03 \\
\cmidrule(lr){2-7}
& XLM-R-Large             & 88.1 & 90.17 & 96.98 & 86.43 & 90.42 \\
& XLM-R-Base              & 87.2 & 88.73 & 96.48 & 85.54 & 89.49 \\
& mMiniLM-L6-H384         & 83.8 & 78.64 & 94.33 & 80.69 & 84.37 \\
& mMiniLM-L12-H384        & 85.0 & 81.54 & 94.84 & 83.02 & 86.10 \\
& de-hplt-og              & 89.5 & 88.45 & 94.85 & 84.56 & 89.34 \\
\midrule

\multirow{11}{*}{\rotatebox[origin=c]{90}{English (en)}}
& TiME-xs                 & 76.5 & 93.25 & 96.23 & 84.15 & 87.53 \\
& TiME-m                  & 80.9 & 95.94 & 97.30 & 90.27 & 91.10 \\
& en-hplt-xs              & 75.7 & 93.64 & 96.79 & 81.93 & 87.02 \\
& en-hplt-m               & 79.0 & 95.72 & 97.44 & 88.42 & 90.14 \\
\cmidrule(lr){2-7}
& XLM-R-Large             & 83.0 & 96.97 & 97.71 & 92.06 & 92.44 \\
& XLM-R-Base              & 81.5 & 96.31 & 97.44 & 90.69 & 91.48 \\
& mMiniLM-L6-H384         & 79.2 & 93.17 & 96.40 & 85.34 & 88.53 \\
& mMiniLM-L12-H384        & 81.3 & 94.39 & 96.65 & 87.75 & 90.02 \\
& en-MiniLM-L6-H768       & 81.5 & 95.72 & 96.95 & 89.44 & 90.90 \\
& en-hplt-og              & 83.4 & 96.11 & 96.71 & 89.33 & 91.39 \\
\midrule

\multirow{8}{*}{\rotatebox[origin=c]{90}{Spanish (es)}}
& TiME-xs           & 87.2 & 93.30 & 97.37 & 86.14 & 91.00 \\
& TiME-s            & 88.9 & 93.74 & 97.55 & 87.71 & 91.97 \\
& TiME-m            & 88.6 & 94.80 & 98.36 & 89.34 & 92.78 \\
\cmidrule(lr){2-7}
& XLM-R-Large       & 90.2 & 95.75 & 99.10 & 91.98 & 94.26 \\
& XLM-R-Base        & 89.5 & 95.06 & 98.40 & 89.37 & 93.08 \\
& mMiniLM-L6-H384   & 88.4 & 93.20 & 97.45 & 87.09 & 91.54 \\
& mMiniLM-L12-H384  & 89.9 & 94.14 & 97.71 & 88.54 & 92.57 \\
& es-hplt-og        & 90.9 & 95.06 & 97.99 & 89.98 & 93.48 \\
\midrule

\multirow{11}{*}{\rotatebox[origin=c]{90}{French (fr)}}
& TiME-xs                 & 84.4 & 95.79 & 96.82 & 87.27 & 91.07 \\
& TiME-s                  & 86.2 & 96.05 & 96.99 & 88.31 & 91.89 \\
& TiME-m                  & 86.9 & 97.36 & 97.83 & 90.87 & 93.24 \\
& fr-hplt-xs              & 80.9 & 95.83 & 97.39 & 78.66 & 88.20 \\
& fr-hplt-m               & 82.8 & 97.01 & 98.07 & 86.54 & 91.11 \\
\cmidrule(lr){2-7}
& XLM-R-Large             & 89.2 & 97.76 & 98.40 & 93.96 & 94.83 \\
& XLM-R-Base              & 88.8 & 97.41 & 98.14 & 92.40 & 94.19 \\
& mMiniLM-L6-H384         & 86.2 & 94.89 & 96.65 & 87.76 & 91.37 \\
& mMiniLM-L12-H384        & 87.1 & 95.82 & 97.19 & 89.76 & 92.47 \\
& fr-hplt-og              & 89.8 & 97.45 & 96.89 & 91.71 & 93.96 \\
\midrule

\multirow{8}{*}{\rotatebox[origin=c]{90}{Irish (ga)}}
& TiME-s            & 68.9 & 77.50 & 90.62 & 73.51 & 77.63 \\
& TiME-m            & 78.0 & 82.00 & 93.54 & 77.94 & 82.87 \\
& ga-hplt-m         & 73.5 & 83.08 & 93.83 & 70.43 & 80.21 \\
\cmidrule(lr){2-7}
& XLM-R-Large       & 80.5 & 84.27 & 93.69 & 79.36 & 84.45 \\
& XLM-R-Base        & 75.2 & 80.68 & 91.84 & 76.28 & 81.00 \\
& mMiniLM-L6-H384   & 52.8 & 72.39 & 86.22 & 70.00 & 70.35 \\
& mMiniLM-L12-H384  & 61.8 & 75.41 & 87.57 & 72.84 & 74.40 \\
& ga-hplt-og        & 76.8 & 83.94 & 90.23 & 70.39 & 80.34 \\
\midrule
\clearpage
\multirow{8}{*}{\rotatebox[origin=c]{90}{Hindi (hi)}}
& TiME-xs           & 82.9 & 88.05 & 98.55 & 88.75 & 89.56 \\
& TiME-s            & 85.3 & 89.13 & 98.63 & 89.65 & 90.68 \\
& TiME-m            & 87.9 & 91.77 & 98.78 & 91.69 & 92.53 \\
\cmidrule(lr){2-7}
& XLM-R-Large       & 87.5 & 92.93 & 98.83 & 92.53 & 92.95 \\
& XLM-R-Base        & 87.5 & 92.12 & 98.76 & 91.57 & 92.49 \\
& mMiniLM-L6-H384   & 81.3 & 86.90 & 98.46 & 88.71 & 88.84 \\
& mMiniLM-L12-H384  & 82.3 & 88.64 & 98.57 & 90.03 & 89.89 \\
& hi-hplt-og        & 89.8 & 92.41 & 98.65 & 91.09 & 92.99 \\
\midrule

\multirow{8}{*}{\rotatebox[origin=c]{90}{Hungarian (hu)}}
& TiME-xs           & 88.0 & 73.01 & 81.15 & 61.97 & 76.03 \\
& TiME-s            & 89.9 & 73.74 & 81.61 & 62.43 & 76.92 \\
& hu-hplt-m         & 88.0 & 82.89 & 86.63 & 52.74 & 77.56 \\
\cmidrule(lr){2-7}
& XLM-R-Large       & 92.6 & 86.27 & 87.69 & 59.91 & 81.62 \\
& XLM-R-Base        & 91.5 & 84.15 & 86.47 & 57.92 & 80.01 \\
& mMiniLM-L6-H384   & 88.6 & 71.92 & 80.93 & 58.89 & 75.09 \\
& mMiniLM-L12-H384  & 89.5 & 73.88 & 82.24 & 61.12 & 76.68 \\
& hu-hplt-og        & 93.2 & 77.19 & 82.73 & 30.43 & 70.89 \\
\midrule

\multirow{8}{*}{\rotatebox[origin=c]{90}{Italian (it)}}
& TiME-xs           & 86.4 & 96.22 & 96.34 & 88.33 & 91.82 \\
& TiME-s            & 87.7 & 96.41 & 96.34 & 89.40 & 92.46 \\
& TiME-m            & 88.6 & 97.58 & 97.89 & 92.71 & 94.20 \\
\cmidrule(lr){2-7}
& XLM-R-Large       & 91.4 & 97.89 & 98.18 & 94.01 & 95.37 \\
& XLM-R-Base        & 89.9 & 97.54 & 97.77 & 92.68 & 94.47 \\
& mMiniLM-L6-H384   & 87.0 & 95.44 & 95.99 & 88.68 & 91.78 \\
& mMiniLM-L12-H384  & 88.2 & 96.31 & 96.17 & 90.79 & 92.87 \\
& it-hplt-og        & 90.6 & 97.24 & 96.58 & 91.80 & 94.05 \\
\midrule

\multirow{8}{*}{\rotatebox[origin=c]{90}{Japanese (ja)}}
& TiME-xs           & 56.1 & 89.99 & 95.34 & 85.25 & 81.67 \\
& TiME-s            & 62.3 & 91.63 & 95.94 & 87.80 & 84.42 \\
& TiME-m            & 63.2 & 94.74 & 97.12 & 90.57 & 86.41 \\
\cmidrule(lr){2-7}
& XLM-R-Large       & 66.8 & 96.14 & 97.63 & 92.36 & 88.23 \\
& XLM-R-Base        & 66.4 & 94.69 & 97.20 & 90.43 & 87.18 \\
& mMiniLM-L6-H384   & 60.4 & 90.09 & 95.25 & 86.45 & 83.05 \\
& mMiniLM-L12-H384  & 59.4 & 91.02 & 95.61 & 87.73 & 83.44 \\
& ja-hplt-og        & 63.0 & 93.95 & 96.96 & 88.94 & 85.71 \\
\midrule

\multirow{7}{*}{\rotatebox[origin=c]{90}{Korean (ko)}}
& TiME-xs           & 80.8 & 84.84 & 90.24 & 83.63 & 84.88 \\
& TiME-s            & 83.6 & 85.60 & 90.65 & 84.63 & 86.12 \\
& TiME-m            & 83.6 & 87.41 & 92.34 & 86.44 & 87.45 \\
\cmidrule(lr){2-7}
& XLM-R-Large       & 88.9 & 89.05 & 93.35 & 88.32 & 89.90 \\
& XLM-R-Base        & 86.2 & 87.91 & 92.71 & 86.90 & 88.43 \\
& mMiniLM-L6-H384   & 80.9 & 84.42 & 89.90 & 83.45 & 84.67 \\
& mMiniLM-L12-H384  & 82.6 & 85.22 & 90.45 & 84.71 & 85.74 \\
\midrule

\multirow{8}{*}{\rotatebox[origin=c]{90}{Portuguese (pt)}}
& TiME-xs           & 86.8 & 92.77 & 96.96 & 79.21 & 88.94 \\
& TiME-s            & 87.9 & 93.08 & 97.09 & 80.51 & 89.65 \\
& TiME-m            & 89.7 & 93.59 & 97.67 & 82.46 & 90.86 \\
\cmidrule(lr){2-7}
& XLM-R-Large       & 91.4 & 94.10 & 98.19 & 84.64 & 92.09 \\
& XLM-R-Base        & 90.1 & 93.87 & 97.94 & 83.48 & 91.35 \\
& mMiniLM-L6-H384   & 87.8 & 92.52 & 96.95 & 79.68 & 89.24 \\
& mMiniLM-L12-H384  & 89.5 & 93.07 & 97.06 & 81.37 & 90.25 \\
& pt-hplt-og        & 91.2 & 93.86 & 97.27 & 82.94 & 91.32 \\
\midrule
\clearpage
\multirow{8}{*}{\rotatebox[origin=c]{90}{Russian (ru)}}
& TiME-xs           & 82.6 & 90.13 & 95.32 & 86.98 & 88.76 \\
& TiME-s            & 85.4 & 91.85 & 96.02 & 90.01 & 90.82 \\
& TiME-m            & 83.7 & 93.18 & 97.15 & 91.48 & 91.38 \\
\cmidrule(lr){2-7}
& XLM-R-Large       & 89.2 & 95.04 & 98.25 & 93.85 & 94.08 \\
& XLM-R-Base        & 87.8 & 94.49 & 97.79 & 93.23 & 93.33 \\
& mMiniLM-L6-H384   & 84.7 & 90.28 & 95.40 & 89.24 & 89.91 \\
& mMiniLM-L12-H384  & 85.9 & 90.96 & 95.04 & 90.46 & 90.59 \\
& ru-hplt-og        & 89.3 & 94.77 & 97.42 & 93.22 & 93.68 \\
\midrule

\multirow{10}{*}{\rotatebox[origin=c]{90}{Urdu (ur)}}
& TiME-xs           & 93.0 & 78.52 & 96.60 & 75.91 & 86.00 \\
& TiME-s            & 92.9 & 78.41 & 96.26 & 77.01 & 86.14 \\
& TiME-m            & 95.3 & 80.16 & 96.99 & 79.91 & 88.09 \\
& ur-hplt-xs        & 87.1 & 62.15 & 92.83 & 54.88 & 74.24 \\
& ur-hplt-m         & 87.9 & 69.97 & 95.00 & 63.35 & 79.06 \\
\cmidrule(lr){2-7}
& XLM-R-Large       & 95.1 & 80.80 & 97.01 & 81.97 & 88.72 \\
& XLM-R-Base        & 94.9 & 79.79 & 96.72 & 79.99 & 87.85 \\
& mMiniLM-L6-H384   & 91.9 & 77.53 & 96.17 & 75.72 & 85.33 \\
& mMiniLM-L12-H384  & 93.2 & 78.49 & 96.35 & 77.73 & 86.44 \\
& ur-hplt-og        & 90.4 & 63.06 & 92.58 & 53.13 & 74.79 \\
\midrule

\multirow{8}{*}{\rotatebox[origin=c]{90}{Chinese (zh)}}
& TiME-xs           & 68.4 & 91.23 & 99.83 & 66.72 & 81.55 \\
& TiME-s            & 70.8 & 91.65 & 99.83 & 68.26 & 82.63 \\
& TiME-m            & 73.0 & 94.31 & 99.89 & 76.18 & 85.84 \\
\cmidrule(lr){2-7}
& XLM-R-Large       & 75.5 & 95.86 & 99.91 & 80.70 & 87.99 \\
& XLM-R-Base        & 76.3 & 95.01 & 99.88 & 76.49 & 86.92 \\
& mMiniLM-L6-H384   & 68.1 & 91.57 & 99.84 & 67.08 & 81.65 \\
& mMiniLM-L12-H384  & 69.9 & 92.43 & 99.84 & 70.15 & 83.08 \\
& zh-hplt-og        & 75.0 & 92.82 & 99.81 & 61.99 & 82.41 \\
\end{longtable}


\begin{table*}[!htbp]
\centering
\resizebox{1.0\linewidth}{!}{%
\renewcommand{\arraystretch}{1.4}
\setlength{\tabcolsep}{2pt} 
\begin{tabular}{@{}l cc cccccccccccccccc ccc@{}}
\toprule
 & & & \multicolumn{16}{c}{\textbf{Average NLP Score per Language}} & & & \\
 \cmidrule(lr){4-19}
\textbf{Model} & \textbf{\shortstack{Params\\(M)}} & \textbf{\#L} & \rotatebox{90}{ar} & \rotatebox{90}{da} & \rotatebox{90}{de} & \rotatebox{90}{en} & \rotatebox{90}{es} & \rotatebox{90}{fr} & \rotatebox{90}{ga} & \rotatebox{90}{hi} & \rotatebox{90}{hu} & \rotatebox{90}{it} & \rotatebox{90}{ja} & \rotatebox{90}{ko} & \rotatebox{90}{pt} & \rotatebox{90}{ru} & \rotatebox{90}{ur} & \rotatebox{90}{zh} & \textbf{\shortstack{Avg\\(16)}} & \textbf{\shortstack{Latency\\Impr.($\times$)}} & \textbf{\shortstack{Throughput\\Impr.($\times$)}} \\
\midrule
\multicolumn{22}{@{}l}{\textit{Baselines}} \\
HPLT                & 150 & 12 & --   & 88.8 & 89.3 & 91.4 & 93.5 & 94.0 & 80.3 & 93.0 & 70.9 & 94.0 & 85.7 & --   & 91.3 & 93.7 & 74.8 & 82.4 & 87.4 & 0.9 & 2.5 \\
XLM-R-Base          & 278 & 12 & 86.5 & 91.0 & 89.5 & 91.5 & 93.1 & 94.2 & 81.0 & 92.5 & 80.0 & 94.5 & 87.2 & 88.4 & 91.3 & 93.3 & 87.8 & 86.9 & 89.3 & 2.0 & 3.2 \\
mMiniLM-L6-H384     & 107 & 6  & 80.4 & 86.2 & 84.4 & 88.5 & 91.5 & 91.4 & 70.3 & 88.8 & 75.1 & 91.8 & 83.0 & 84.7 & 89.2 & 89.9 & 85.3 & 81.7 & 85.1 & 3.7 & 15.4 \\
XLM-R-Large         & 560 & 24 & 88.7 & 92.5 & 90.4 & 92.4 & 94.3 & 94.8 & 84.5 & 93.0 & 81.6 & 95.4 & 88.2 & 89.9 & 92.1 & 94.1 & 88.7 & 88.0 & 90.5 & 1.0 & 1.0 \\
\midrule
\multicolumn{22}{@{}l}{\textit{Our Models (TiME)}} \\
TiME-*-m     & 236 & 6  & 86.4 & 90.2 & 88.7 & 91.1 & 92.8 & 93.2 & 82.9 & 92.5 & 81.25 & 94.2 & 86.4 & 87.5 & 90.9 & 91.4 & 88.1 & 85.8 & 89.5 & 3.7 & 6.3 \\
TiME-*-s              & 107 & 6  & 82.6 & 87.6 & 86.7 & 88.9 & 92.0 & 91.9 & 77.6 & 90.7 & 76.9 & 92.5 & 84.4 & 86.1 & 89.7 & 90.8 & 86.1 & 82.6 & 86.6 & 3.9 & 15.3 \\
TiME-*-xs             & 103 & 4  & 81.2 & 86.2 & 84.8 & 87.5 & 91.0 & 91.1 & 75.5   & 89.6 & 76.0 & 91.8 & 81.7 & 84.9 & 88.9 & 88.8 & 86.0 & 81.5 & 86.1 & 5.4 & 23.2 \\
*-hplt-m            & 69  & 6  & --   & 88.1 & 87.0 & 90.1 & --   & 91.1 & 80.2 & --   & 77.6 & --   & --   & --   & --   & --   & 79.1 & --   & 84.8 & 3.7 & 6.6 \\
\bottomrule
\end{tabular}
}
\caption{\textbf{Complete summary of average NLP task scores and efficiency metrics across all 16 evaluated languages.} The `Latency Improvement` and `Throughput Improvement` metrics are calculated relative to XLM-R-Large. Empty cells (`--`) indicate that data for a specific model-language combination was not available. The '*' is a placeholder for the language, and *-hplt-m refers to the models that were distilled from the HPLT model as a Teacher.}
\label{tab:appendix_full_summary_16_langs}
\end{table*}

\section{Downstream Task Datasets}
\label{apx:datasets_detailed}

Our evaluation relies on established benchmark datasets for each NLP task. For part-of-speech (POS) tagging, lemmatization, and dependency parsing (LAS), we use specific treebanks from Universal Dependencies (UD) \citep{de2021universal}, primarily following the selections in the HPLT evaluation framework \citep{pyysalo2024hplt}. For named entity recognition (NER), we predominantly use language-specific splits from the WikiAnn dataset \citep{rahimi-etal-2019-massively, pan-etal-2017-cross}, unless a more standard dataset is conventionally used for a specific language. The details are as follows:

\subsection{Arabic (ar)}
\begin{itemize}
    \item \textbf{POS, Lemma, LAS:} UD Arabic-PADT \cite{hajic2004padt, ud_arabic_padt}
    \item \textbf{NER:} WikiAnn Arabic split \cite{rahimi-etal-2019-massively}.
\end{itemize}

\subsection{Chinese (zh)}
\begin{itemize}
    \item \textbf{POS, Lemma, LAS:} UD Chinese-GSD \cite{ud_chinese_gsd}
    \item \textbf{NER:} WikiAnn Chinese split \cite{rahimi-etal-2019-massively}.
\end{itemize}

\subsection{Danish (da)}
\begin{itemize}
    \item \textbf{POS, Lemma, LAS:} UD Danish-DDT \cite{johannsen2015universal} 
    \item \textbf{NER:} WikiAnn Danish split \cite{rahimi-etal-2019-massively}. 
\end{itemize}

\subsection{German (de)}
\begin{itemize}
    \item \textbf{POS, Lemma, LAS:} UD German-GSD \cite{mcdonald2013universal, borges-volker-etal-2019-hdt} 
    \item \textbf{NER:} WikiAnn German split \cite{rahimi-etal-2019-massively}. 
\end{itemize}

\subsection{English (en)}
\begin{itemize}
    \item \textbf{POS, Lemma, LAS:} UD English-EWT \cite{silveira-etal-2014-gold}
    \item \textbf{NER:} WikiAnn English split \cite{rahimi-etal-2019-massively}. 
\end{itemize}

\subsection{Spanish (es)}
\begin{itemize}
    \item \textbf{POS, Lemma, LAS:} UD Spanish-AnCora \cite{taule2008ancora, ud_spanish_ancora}
    \item \textbf{NER:} WikiAnn Spanish split \cite{rahimi-etal-2019-massively}.
\end{itemize}

\subsection{French (fr)} 
\begin{itemize}
    \item \textbf{POS, Lemma, LAS:} UD French-GSD \cite{guillaume:hal-02267418, mcdonald2013universal} 
    \item \textbf{NER:} WikiAnn French split \cite{rahimi-etal-2019-massively}.
\end{itemize}

\subsection{Irish (ga)}
\begin{itemize}
    \item \textbf{POS, Lemma, LAS:} UD Irish-IDT \cite{lynn2016universal, ud_irish_idt}
    \item \textbf{NER:} WikiAnn Irish split \cite{rahimi-etal-2019-massively}.
\end{itemize}

\subsection{Hindi (hi)}
\begin{itemize}
    \item \textbf{POS, Lemma, LAS:} UD Hindi-HDTB \cite{bhatt2009hindiurdu, ud_hindi_hdtb}
    \item \textbf{NER:} WikiAnn Hindi split \cite{rahimi-etal-2019-massively}.
\end{itemize}

\subsection{Hungarian (hu)}
\begin{itemize}
    \item \textbf{POS, Lemma, LAS:} UD Hungarian-Szeged \cite{vincze2010hungarian, ud_hungarian_szeged}
    \item \textbf{NER:} WikiAnn Hungarian split \cite{rahimi-etal-2019-massively}.
\end{itemize}

\subsection{Italian (it)}
\begin{itemize}
    \item \textbf{POS, Lemma, LAS:} UD Italian-ISDT \cite{bosco2014evalita, ud_italian_isdt}
    \item \textbf{NER:} WikiAnn Italian split \cite{rahimi-etal-2019-massively}.
\end{itemize}

\subsection{Japanese (ja)}
\begin{itemize}
    \item \textbf{POS, Lemma, LAS:} UD Japanese-GSD \cite{ud_japanese_gsd}
    \item \textbf{NER:} WikiAnn Japanese split \cite{rahimi-etal-2019-massively}.
\end{itemize}

\subsection{Korean (ko)}
\begin{itemize}
    \item \textbf{POS, Lemma, LAS:} UD Korean-Kaist \cite{chun2018udkorean, ud_korean_kaist}
    \item \textbf{NER:} WikiAnn Korean split \cite{rahimi-etal-2019-massively}.
\end{itemize}

\subsection{Portuguese (pt)}
\begin{itemize}
    \item \textbf{POS, Lemma, LAS:} UD Portuguese-CINTIL \cite{branco2012cintil, ud_portuguese_cintil}
    \item \textbf{NER:} WikiAnn Portuguese split \cite{rahimi-etal-2019-massively}.
\end{itemize}

\subsection{Russian (ru)}
\begin{itemize}
    \item \textbf{POS, Lemma, LAS:} UD Russian-SynTagRus \cite{boguslavsky2000syntagrus, ud_russian_syntagrus}
    \item \textbf{NER:} WikiAnn Russian split \cite{rahimi-etal-2019-massively}.
\end{itemize}

\subsection{Urdu (ur)}
\begin{itemize}
    \item \textbf{POS, Lemma, LAS:} UD Urdu-UDTB \cite{bhathindi, ud_urdu_udtb, palmer2009hindi}
    \item \textbf{NER:} WikiAnn Urdu split \cite{rahimi-etal-2019-massively}.
\end{itemize}

For question answering tasks in English and German, the \textbf{MLQA} dataset was used \cite{lewis-etal-2020-mlqa}.
The primary distillation corpus for all languages is CulturaX \citep{nguyen2023culturaxcleanedenormousmultilingual}.
Checkpoint selection relies on development splits from FLORES-200 \citep{goyal2022flores101, flores_hf} for Irish, and WMT24++ \citep{wmt24pp, wmt24pp_arxiv} for other languages.

\end{document}